%% file: root.tex
\documentclass[letterpaper,10pt,conference]{ieeeconf}

\IEEEoverridecommandlockouts
\usepackage{amsmath}
\usepackage{amssymb}
\usepackage{mathrsfs}
\usepackage{bm}

\usepackage{graphicx}
\usepackage{booktabs}
\usepackage{multirow}
\usepackage{makecell}
\usepackage{capt-of}
\usepackage{array}
\usepackage{tabularx}
\usepackage{listings}
\usepackage{float}
\usepackage{placeins}
\usepackage{stfloats}

\usepackage{pifont}
\usepackage{xspace}
\usepackage{xcolor}
\usepackage[english]{babel}
\usepackage{cite}
\usepackage[hidelinks]{hyperref}

\newcommand{\ours}{HINT}
\newcommand{\pis}{\ensuremath{\pi_{0.5}}}
\newcommand{\subsr}{Sub.\ SR}
\newcommand{\fullsr}{Full SR}

\newcolumntype{Y}{>{\raggedright\arraybackslash}X}
\definecolor{lightblue}{RGB}{80, 160, 220}
\lstdefinestyle{prompt}{
  basicstyle=\ttfamily\scriptsize,
  columns=fullflexible,
  keepspaces=true,
  breaklines=true,
  frame=single,
  framerule=0.25pt,
  rulecolor=\color{black!35},
  xleftmargin=0.4em,
  xrightmargin=0.4em,
  aboveskip=0.35em,
  belowskip=0.35em,
  showstringspaces=false
}

\newcounter{mainsectioncheckpoint}
\newcounter{mainfigurecheckpoint}
\newcounter{maintablecheckpoint}
\newcounter{mainequationcheckpoint}

\providecommand{\IEEEPARstart}[2]{#1#2}

\makeatletter

\newcommand{\Rmnum}[1]{%
    \expandafter\@slowromancap\romannumeral #1@}
\makeatother

\title{\LARGE \bf
HINT: Human-Intent Inception for Agentic Robot Manipulation
}

\author{
\authorblockN{
Mingyu Mei$^{1}$, Haojie Xu$^{1}$, Shihao Jin$^{1}$, Zibo Dai$^{1}$, Qihao Cheng$^{1}$, Zhengrui Lv$^{1}$,
}
\authorblockN{
Hongjie Fang$^{2}$, Shirun Tang$^{3}$, Guang Chen$^{1,4}$, Xinyue Zhao$^{1}$, Huiliang Shen$^{1}$, Zaixing He$^{1,\dagger}$
}
\authorblockA{
$^{1}$Zhejiang University\quad
$^{2}$Shanghai Jiao Tong University\quad
$^{3}$Noematrix\quad
$^{4}$EndlessAI\quad
}
\authorblockA{
$^{\dagger}$Corresponding author
}
\authorblockA{
\href{mailto:mingyumei@zju.edu.cn}{\texttt{mingyumei@zju.edu.cn}},
\quad
\href{mailto:zaixinghe@zju.edu.cn}{\texttt{zaixinghe@zju.edu.cn}}
}
}

\begin{document}

\IEEEaftertitletext{%
    \begin{center}
    \centering
    \includegraphics[width=0.98\textwidth]{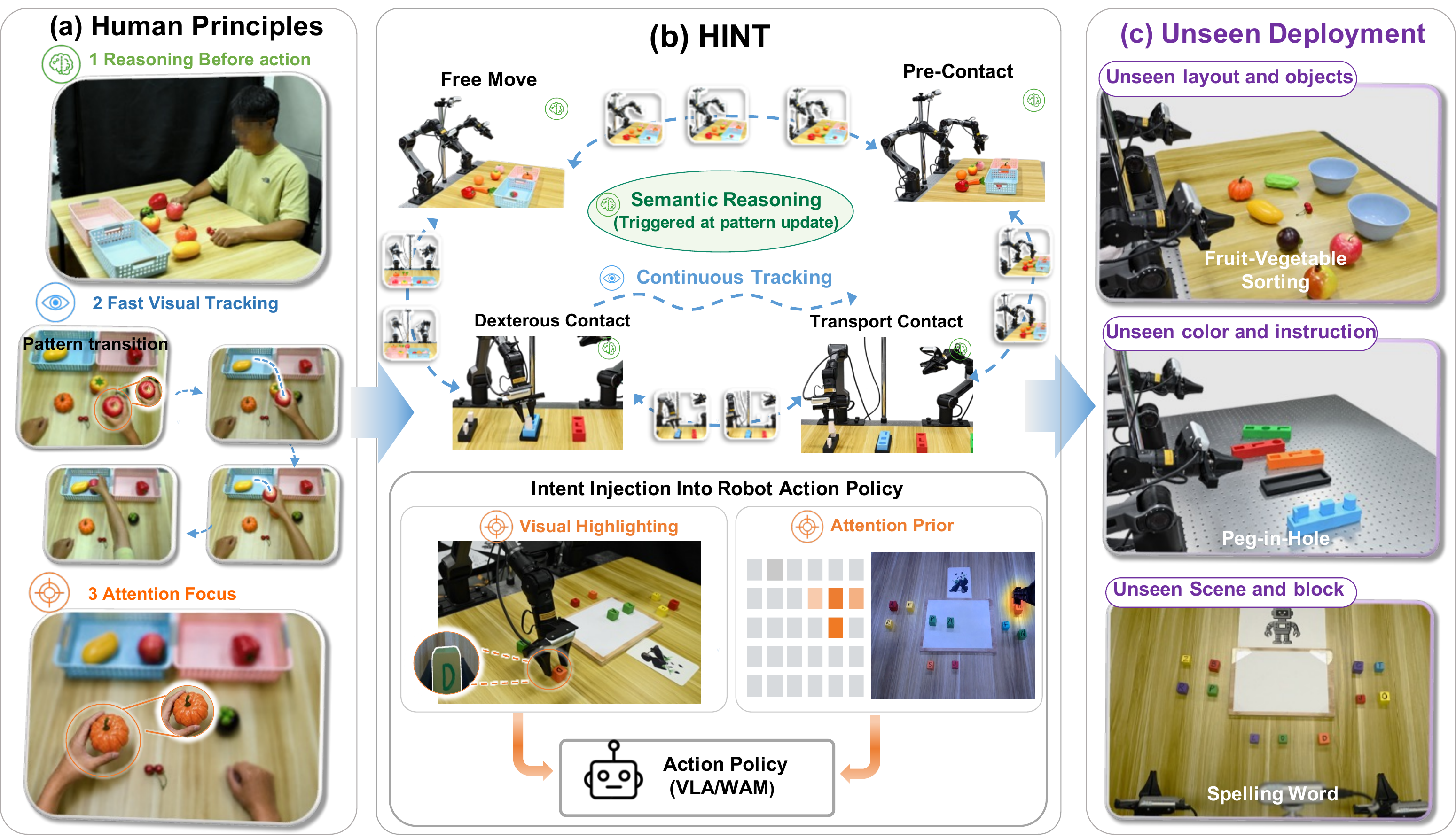}
    \captionof{figure}{Motivation and overview of \textbf{HINT} (\textbf{H}uman-\textbf{INT}ent \textbf{IN}cep\textbf{T}ion). (a) Human manipulation strategies that inspire HINT: humans identify the intended target before acting, track it throughout motion, and maintain target-focused attention during interaction. (b) HINT translates these strategies into semantic reasoning at manipulation-pattern transitions, continuous target tracking between transitions, and spatial-intent injection into the action policy via visual highlighting and an attention prior. (c) HINT is evaluated under unseen semantic-compositional variations in objects, layouts, colors, and instructions.}
    \label{fig:teaser}
    \end{center}
}

\maketitle
\thispagestyle{empty}
\pagestyle{empty}

\input{0_abstract}
\input{1_introduction}
\input{2_related}

\input{3_method}
\input{4_experiment}
\input{5_conclusion}
\input{6_limitations}

\bibliographystyle{ieeetr}
\bibliography{ref}

\setcounter{mainsectioncheckpoint}{\value{section}}
\setcounter{mainfigurecheckpoint}{\value{figure}}
\setcounter{maintablecheckpoint}{\value{table}}
\setcounter{mainequationcheckpoint}{\value{equation}}
\clearpage
\begingroup
\setcounter{section}{0}
\setcounter{figure}{0}
\setcounter{table}{0}
\setcounter{equation}{0}
\renewcommand{\thesection}{Supp.\ \Roman{section}}
\renewcommand{\thesubsection}{\thesection.\arabic{subsection}}
\renewcommand{\thefigure}{S\arabic{figure}}
\renewcommand{\thetable}{S\arabic{table}}
\renewcommand{\theequation}{S\arabic{equation}}
\renewcommand{\theHsection}{supp.\arabic{section}}
\renewcommand{\theHsubsection}{supp.\arabic{section}.\arabic{subsection}}
\renewcommand{\theHfigure}{supp.\arabic{figure}}
\renewcommand{\theHtable}{supp.\arabic{table}}
\renewcommand{\theHequation}{supp.\arabic{equation}}
\section*{Supplementary Materials for HINT}
\input{supplementary}

\clearpage
\endgroup
\setcounter{section}{\value{mainsectioncheckpoint}}
\setcounter{figure}{\value{mainfigurecheckpoint}}
\setcounter{table}{\value{maintablecheckpoint}}
\setcounter{equation}{\value{mainequationcheckpoint}}

\end{document}

%% file: 0_abstract.tex
\begin{abstract}
Humans can perform complex manipulations given a simple intent through an overall instruction, while continuously adapting to evolving visual observations. However, current vision-language action (VLA) models and other action policies struggle to realize this high-level intelligent behavior under dense, evolving visual inputs and sparse language guidance. Visual correlations can then dominate semantic intent, leading actions to follow visual shortcuts rather than human goals. We present HINT (Human-INTent INcepTion), an agentic framework inspired by the human manipulation principles: semantic intent changes sparsely at manipulation-pattern transitions, whereas continuous control primarily depends on the evolving object-hand relationship. HINT invokes semantic reasoning only at pattern transitions to resolve the current subtask and target, then maintains this commitment through multi-view grounding and visual tracking. We explore two visual interfaces-image-space semantic highlighting and attention-prior injection-to communicate the tracked intent to the action policy without introducing additional trainable parameters into the foundation action model. Experiments across three long-horizon tasks and out-of-distribution variants show that HINT substantially improves intent understanding, task progress, and end-to-end success across two foundation policies while preserving low-latency control. Project website: \color{lightblue}{https://robot-hint.github.io/}.
\end{abstract}


%% file: 1_introduction.tex
\section{INTRODUCTION}
\noindent\hfill
\begin{minipage}{0.78\linewidth}
\centering

{\small\itshape
A simple little thought that changes everything\ldots
\par}

\vspace{0.35em}
\hrule
\vspace{0.25em}

\hfill
{\small Christopher Nolan, \textit{Inception}}

\end{minipage}
\par
\vspace{0.5\baselineskip}

\IEEEPARstart{R}{ecent} advances in imitation learning~\cite{diffusion_policy,act,dp3,rise}, Vision-Language-Action (VLA) models~\cite{kim2024openvla,black2024pi0,black2025pi05}, and World Action Models~\cite{dreamzero,cosmos_policy} have substantially improved generalization and closed-loop control in robotic manipulation. Combined with DAgger and reinforcement learning~\cite{dagger,rl_token,rl100}, these models achieve strong performance within constrained task distributions. However, end-to-end foundation models face an inherent tension between semantic understanding and action control: adapting representations to robot dynamics will weaken pretrained semantic capabilities, while static language instructions and evolving visual observations encourage visual shortcuts and intent drift. This limitation is amplified in long-horizon manipulation, thereby introducing a fundamental challenge: how to seamlessly couple sparse semantic reasoning with continuous visuomotor feedback to enable robust execution without incurring prohibitive inference latency.

Prior work has integrated semantic reasoning into VLA models from several complementary perspectives. Some methods enhance action-relevant perception through additional VLM reasoning or selectively alternate between reasoning and acting to reduce inference cost~\cite{zhang2026uam,lin2025onetwovla}. Others inject spatial guidance through masks, points, boxes, or trajectories~\cite{yang2026lilovla,wang2026vpvla,liu2026lohomanip,liang2025pixelvla}, or adapt visual observations to improve task-relevant visibility~\cite{fan2025longvla,liu2026activevla}. However, these methods remain limited in supporting continuous semantic understanding over long-horizon tasks. Consequently, action policies learn semantic variations from demonstrations rather than reason through a general semantic capability, leading to high data demands and limited generalization to unseen task compositions.

Human manipulation is not uniformly semantic over time. High-level goals remain stable, whereas low-level movements are continuously corrected through sensorimotor feedback~\cite{uithol2012hierarchies,desmurget2000forward,todorov2002optimal}. Visual attention is likewise directed toward objects relevant to the current or upcoming action~\cite{land2001eye,hayhoe2005eye}. As illustrated in Fig.~\ref{fig:teaser}(a), humans identify the intended target before acting, track how the hand and objects move within a manipulation pattern, and maintain attention on the target-hand interaction. These observations suggest a simple temporal principle: semantic reasoning is mainly needed when the manipulation pattern changes; within a pattern, execution primarily requires maintaining where the target is relative to the hand. Human intent therefore changes sparsely, but its visual realization must remain continuous.

Motivated by this principle, we propose \textbf{HINT} (\textbf{H}uman-\textbf{INT}ent \textbf{IN}cep\textbf{T}ion), an agentic framework that turns sparse language intent into continuous, target-specific visual guidance. HINT closes the semantic control loop in three steps. First, \emph{Pattern-Aware Perception Scheduling} detects the current manipulation pattern, selects the informative camera views, and determines when semantic reasoning is required. Second, \emph{The Semantic Commitment and Visual Tracking} resolves the current subtask and target at pattern transitions, then continuously tracks that target until the next transition. Third, \emph{View-Routed Semantic Intent Injection} communicates the tracked target to the action policy through image-level visual highlighting and token-level attention bias. This decomposition separates \emph{what} to act on from \emph{how} to act: HINT maintains human intent, while the pretrained action policy retains responsibility for motor execution. The complete system is summarized in Fig.~\ref{fig:framework}. Our contributions are threefold:

\begin{itemize}
    \item \textbf{We propose HINT, a framework that translates high-level intent
    into long-horizon robot manipulation} by converting sparse
    language instructions into dense, continuous semantic visual cues across
    multiple views.

    \item \textbf{We introduce pattern-aware semantic perception scheduling,}
    which uses manipulation-pattern transitions to trigger semantic reasoning
    and multi-view routing. By coupling sparse intent updates with continuous
    visuomotor control, this mechanism invokes semantic perception only at task-relevant transitions, reducing latency while maintaining responsive execution.

    \item \textbf{We introduce a parameter-free semantic interface} that
    continuously communicates semantic intent to foundation action policies
    through complementary image-level highlighting and attention-level
    guidance, without adding trainable parameters to the action backbone.
\end{itemize}

Across three tasks and both ID and semantic-compositional OOD settings, HINT
consistently improves two foundation action policies while retaining
low-latency control. Averaged over the six task--setting pairs, it increases
Intention Score (IS), Subtask Success Rate (Sub.\ SR), and Full-Task Success
Rate (Full SR) by \textbf{32.9/29.0/18.6} percentage points on
Wall-OSS-0.5~\cite{yu2026walloss05technicalreport}, respectively and by \textbf{50.2/41.1/41.4}  points on
$\pi_{0.5}$~\cite{black2025pi05}.

%% file: 2_related.tex
\section{Related Work}
\label{sec:Related_Work}

\subsection{Long-Horizon Manipulation and Interaction Patterns}

Long-horizon manipulation combines high-level task organization with reliable execution across multiple interactions. Prior work predicts semantic subtasks~\cite{black2025pi05}, alternates reasoning and control~\cite{shi2025hirobot,lin2025onetwovla}, updates latent plans at low frequency~\cite{huang2025thinkact}, composes object-centric policies~\cite{yang2026lilovla}, or uses visual trajectories, affordances, and progress estimates to organize execution~\cite{liu2026lohomanip,liu2026palm}. VLABench evaluates such capabilities through multi-stage tasks requiring commonsense and implicit-intent reasoning~\cite{vlabenchmark}. These approaches largely decide which subtask, skill, or policy should execute next, while perception often remains unchanged within a subtask.

Long-VLA and BFA both rely on predefined manipulation phases to adaptively prioritize global and wrist-view information~\cite{fan2025longvla,BFA}. See Selectively, Act Adaptively similarly exploits interaction structure to route wrist observations and action experts for bimanual manipulation~\cite{choi2026selective}, whereas other stage-aware methods regulate progress estimation, reactive transitions, or force control~\cite{feng2024playtoscore,he2024foar,li2026forcevla2}.

For long-horizon manipulation, prior methods often rely on task-specific subtasks or predefined stages, limiting reuse across interaction structures. In contrast, HINT uses a shared four-pattern vocabulary that is independent of task identity and jointly determines when semantic reasoning is invoked and which views or sensors should be prioritized. In our experiments, a single Pattern Router is trained jointly on all three task datasets, providing a unified schedule for reasoning and perception throughout execution.

\subsection{Reasoning for Robotic Manipulation}

Generalist vision-language-action (VLA) policies inherit broad semantic priors
from vision-language pretraining, enabling multi-task robot control~\cite{kim2024openvla,octo2024,black2024pi0,black2025pi05,
gemini2025robotics15,zhou2025chatvla}. Long-horizon manipulation, however,
exposes a structural mismatch: language specifies abstract intent, whereas
control requires that intent to remain spatially grounded as the scene evolves.
Existing methods bridge this gap through intermediate representations, ranging
from language subtasks, grounded plans, and motion primitives to visual traces,
hierarchical spatial decisions, and latent visual
plans~\cite{belkhale2024rth,zawalski2024ecot,zhou2025chatvla,zhao2025cotvla,
li2025hamster,huang2025thinkact,chen2026dial,yang2026hivla}. These
representations improve semantic-action alignment, but richer reasoning paths
can substantially increase inference cost, as illustrated by multi-expert
architectures such as UAM~\cite{zhang2026uam}.

To reduce reasoning overhead, OneTwoVLA adaptively switches between action and reasoning modes, reasoning only upon errors or subtask completion~\cite{lin2025onetwovla}. Other existing work uses execution memory and visual foresight~\cite{shou2026halo,koo2026hamlet}, or compact latent reasoning~\cite{zhong2026acotvla,huang2026fastthinkact,li2026trmvla,bai2026laravla,wu2026continuous}. Feedback-aware systems additionally revise plans based on execution outcomes, motion feasibility, or failures~\cite{shah2025bumble,wang2024llm3}. 

HINT builds on this insight: semantic intent changes sparsely, whereas its visual realization changes continuously. Manipulation-pattern transitions determine when to jointly resolve the current subtask and localize its target from the routed views; tracking then maintains the commitment through continuous scene evolution. HINT therefore preserves persistent intent without repeatedly reinterpreting language, aligning semantic reasoning with execution dynamics while reducing redundant inference and latency.

\subsection{Visual Grounding for Manipulation}

Implicit grounding improves task-relevant visual representations without altering policy observations. ReconVLA reconstructs manipulation-relevant gaze regions as an auxiliary objective~\cite{song2025reconvla}. Other representative approaches enhance perception through 3D alignment, affordance modeling, hierarchical grounding, and history-conditioned attention~\cite{li2025spatialforcing,huang2025otter,yu2026affordancevla,chen2026dial,yang2026hivla,zhou2025chatvla,xiao2026avavla}. These methods strengthen visual features, but typically retain a fixed perception throughout execution.

Explicit grounding, by contrast, communicates semantic intent directly through the visual input. VP-VLA overlays planner-generated points and boxes on policy observations~\cite{wang2026vpvla}, while LiLo-VLA suppresses irrelevant content through object-centric masks~\cite{yang2026lilovla}. Other methods provide visual trajectories~\cite{liu2026lohomanip}, jointly predict actions and future masks~\cite{yu2026maskwam}, or ground manipulation through keypoints, candidate renderings, motion traces, and projected end-effector cues~\cite{fang2024moka,nasiriany2024pivot,zheng2024tracevla,dai2025aimbot}. Pixel-level prompts~\cite{liang2025pixelvla}, active viewpoint selection~\cite{liu2026activevla}, and reasoning-derived visual aids~\cite{yu2025point,yuan2025fsd,zhao2026lsr} provide further forms of spatial guidance.

In contrast, HINT does not learn an additional grounding pathway; it makes the resolved intent explicit to the action policy. Highlighting externalizes the intended target, while the attention prior keeps policy focus on the object during action generation. These intent-injection interfaces introduce no additional trainable parameters into the foundation action policy, enabling the same semantic interface to be applied across different policy backbones.

%% file: 3_method.tex
\section{Method}

\begin{figure*}[t!]
    \centering
    \vspace{1.5em}
    \includegraphics[width=1.0\linewidth]{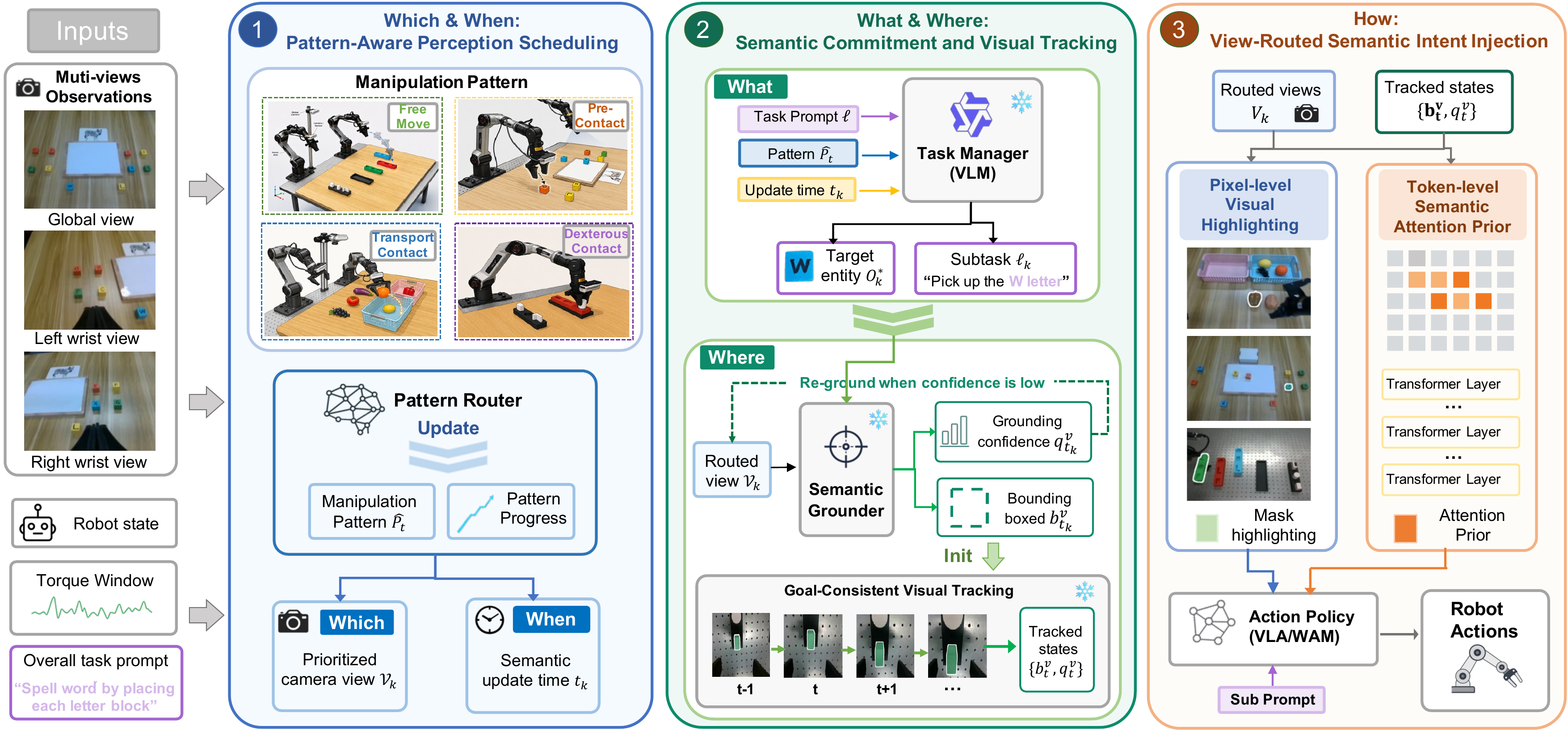}
    \caption{System overview of HINT. HINT consists of three parts: (1) The pattern router estimates the current manipulation pattern
    and its progress to select the prioritized camera views $\mathcal{V}_k$ and
    the semantic update time $t_k$. (2) At an update, the task manager resolves
    the active subtask and target entity, the semantic grounder localizes the
    target in the routed views, and goal-consistent tracking maintains the
    target state between updates;  (3) The
    intent is injected into the action policy through pixel-level
    highlighting and a token-level semantic attention bias to generate robot
    actions.}
    \vspace{-1em}
    \label{fig:framework}
\end{figure*} 

\subsection{Which and When: Pattern-Aware Perception Scheduling}
\label{sec:which and when}

Human manipulation naturally shifts perceptual attention across interaction stages. Before contact, vision primarily guides target localization and alignment. Once contact is established, force and tactile feedback become critical for interaction and fine control. Inspired by this stage-dependent multisensory behavior, we introduce pattern-aware perception scheduling, which uses the current manipulation pattern to determine \textbf{which view or modality to prioritize}, enabling the scheduler to adaptively emphasize the most informative perception sources. The pattern further determines \textbf{when semantic perception should be updated}, providing the temporal schedule for the subsequent perception module.

\textbf{Manipulation Pattern.} We categorize general manipulation behaviors into four manipulation patterns: free move, pre-contact, transport contact, and dexterous contact. The four manipulation patterns define a shared, task-agnostic interface for perception control and cover both pick-and-place and contact-rich manipulation in our evaluated tasks. Each pattern induces a distinct perceptual demand, with different perception sources emphasized accordingly:
\begin{itemize}
\item \textbf{Free move}: The robot moves in free space to approach the target. The \textbf{global view} captures scene-level context for coarse target localization and motion planning.
\item \textbf{Pre-contact}: The robot approaches the target and prepares to grasp or contact. The \textbf{wrist views} offer local geometric cues for precise alignment.
\item \textbf{Transport contact}: The robot maintains contact with the object and transports it toward a target region. The \textbf{global view} provides spatial information about the robot, manipulated object, and goal region.
\item \textbf{Dexterous contact}: The robot performs contact-rich manipulation through sustained interaction. \textbf{Wrist views} provide local visual cues, while \textbf{force/tactile} feedback captures contact dynamics for fine control.
\end{itemize}

\begin{figure}[t!]
    \centering
    \includegraphics[width=1.0\linewidth]{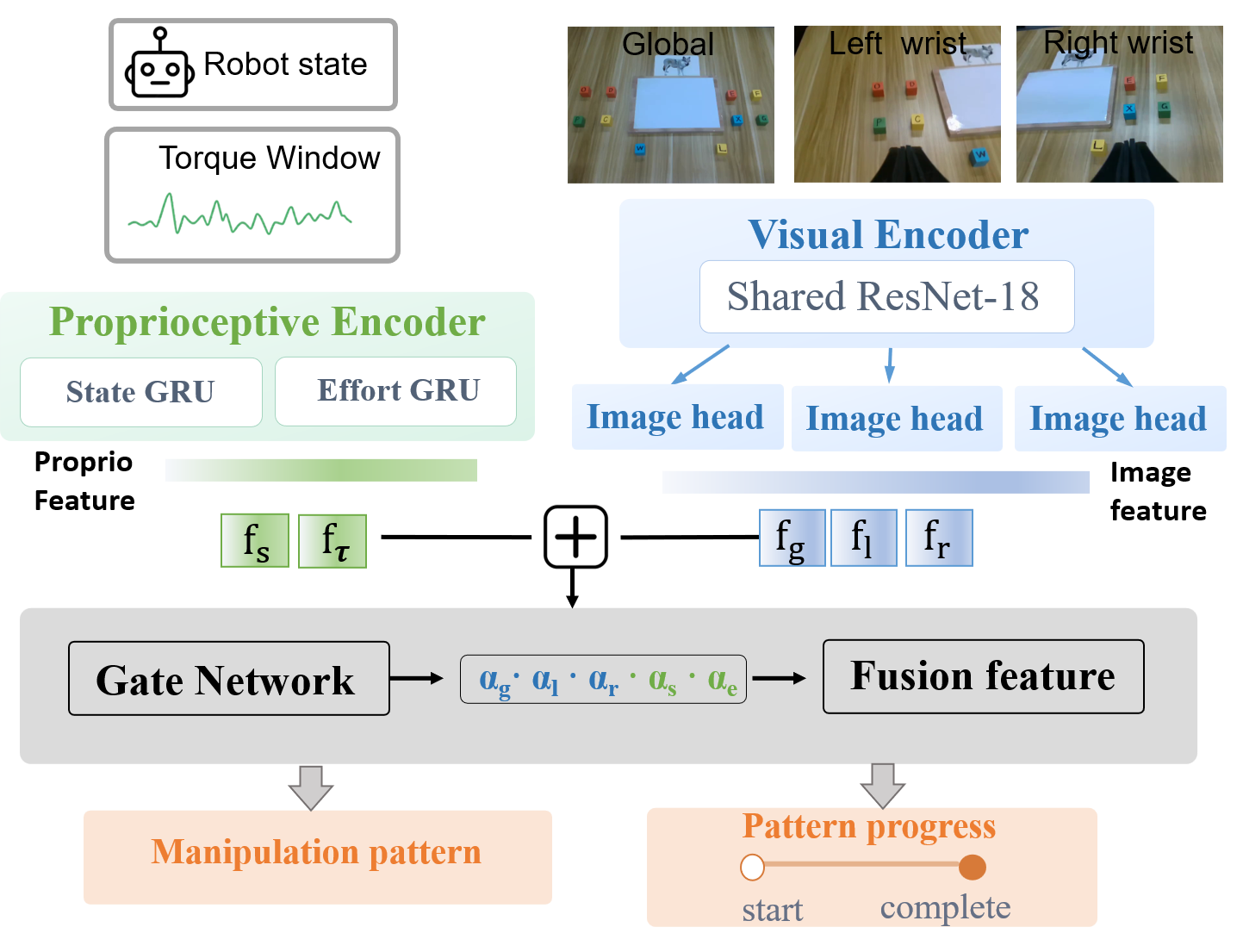}
    \caption{Architecture of the manipulation-pattern router.}
    \label{fig:pattern_network}
    \vspace{-1.5em}
\end{figure}

\textbf{Pattern Router.} To infer the manipulation pattern online, we introduce the \textbf{Pattern Router} (Fig.~\ref{fig:pattern_network}), a lightweight multimodal network conditioned on multi-view visual observations $\mathcal{I}_t = \{I_t^{\text{g}}, I_t^{\text{l}}, I_t^{\text{r}}\}$, proprioceptive state $\mathbf{s}_t^p$, and joint torques $\boldsymbol{\tau}_t$.
The multi-view images are encoded by a shared ResNet-18 backbone~\cite{he2016deep} with view-specific adapters, while the temporal dynamics of proprioception and torques are modeled by separate GRU branches~\cite{cho2014learning}. A gated fusion mechanism adaptively reweights and integrates these features, and the resulting representation is decoded to jointly predict the manipulation pattern $\hat{P}_t$ and the normalized within-pattern progress $\hat{\phi}_t \in [0,1]$.

Crucially, the inferred pattern $\hat{P}_t$ and pattern progress $\hat{\phi}_t \in [0,1]$ determine the prioritized camera view $\mathcal{V}_k \subseteq \{v^g,v^l,v^r\}$ and pattern transition $t_k$ for downstream perception. The network is jointly optimized via:
\begin{equation}
\mathcal{L}_{\mathrm{router}}
=
\mathcal{L}_{\mathrm{CE}}(P_t,\hat{P}_t)
+
0.5*{\mathrm{MSE}}(\phi_t,\hat{\phi}_t),
\end{equation}
where $P_t, \phi_t$ denote the ground-truth manipulation patterns and progress, respectively. We train the Pattern Router jointly on the pattern annotations from all three evaluated tasks.

\subsection{What and Where: Semantic Commitment and Visual Tracking}\label{sec:what and where}
High-level semantic intent changes sparsely, whereas low-level visual attention evolves continuously during manipulation. Humans first determine \textbf{what} to act on by identifying the task-relevant target and its intended role; once this semantic commitment is established, perception primarily tracks \textbf{where} the target is relative to the hand and surrounding scene. Following this perception process, we decompose semantic perception into two perception modes: \textbf{Pattern-Triggered Semantic Reasoning}, which establishes or updates the semantic commitment when the manipulation context changes, and \textbf{Goal-Consistent Visual Tracking}, which continuously maintains the spatial realization of the committed target during execution. This design preserves goal-relevant visual attention with minimal overhead, avoiding redundant reasoning at each inference step.

\paragraph{\textbf{Pattern-Triggered Semantic Reasoning}}
Pattern-Triggered Semantic Reasoning uses the \emph{Task Manager} to resolve the current semantic target from the high-level task instruction. The resulting \emph{semantic commitment} is then localized in the routed camera views by the Semantic Grounder.

\textbf{Task Manager.} Let $\mathcal{T}$ denote the Task Manager and $\ell$ denote the overall task instruction provided by the human, such as “Classify fruits and vegetables into the blue and red baskets, respectively.” At reset, the Task Manager first interprets the overall instruction and scene to establish an ordered subtask plan, which specifies the intended execution sequence. Thereafter, at each pattern transition $t_k$, the VLM-based Task Manager (Qwen3-VL~\cite{bai2025qwen3}) updates the active subtask according to this plan and the current execution context, generating a pattern-conditioned instruction $\ell_k$ (e.g., “pick up the apple”) together with an active target entity $o_k^\star$ (e.g., “orange, carrot, or blue basket”), which collectively define the current subtask and identify the physical elements relevant to its execution. This output is formulated as:

\begin{equation}
(\ell_k, o_k^\star)=
\mathcal{T}
\left(
\ell, \hat{P}_{t_k}, \mathcal{C}_k
\right).
\end{equation}

The generation is conditioned on the overall instruction $\ell$, manipulation pattern $\hat{P}_{t_k}$, and execution context $\mathcal{C}_k$. 
The execution context $\mathcal{C}_k$ is initialized from the reset-time subtask plan and subsequently updated with completed subtasks and the active subtask index. It therefore provides the Task Manager with the accumulated task progress for resolving the next subtask and target entity. The resulting $\ell_k$ and $o_k^\star$ specify \textbf{what} to execute and \textbf{what} to interact with, providing the Semantic Grounder with an unambiguous query for determining \textbf{where} the target appears in visual space.

\textbf{Semantic Grounder.}
Given the resolved subtask $\ell_k$ and target entity $o_k^\star$, the Semantic Grounder $\mathcal{G}$ determines its image-space location in each routed camera view. For $v\in\mathcal{V}_k$, the model predicts:
\begin{equation}
\left(
\mathbf{b}_{t_k}^{v},
q_{t_k}^{v}
\right)
=
\mathcal{G}
\left(
I_{t_k}^{v},
\ell_k,
o_k^\star
\right),
\qquad
v\in\mathcal{V}_k,
\label{eq:semantic_grounding}
\end{equation}
where
$\mathbf{b}_{t_k}^{v}
=
[x_{\min},y_{\min},x_{\max},y_{\max}]^\top$
is the bounding box of $o_k^\star$ in view $v$, and
$q_{t_k}^{v}\in[0,1]$ denotes its grounding confidence.
Specifically, $\mathcal{G}$ leverages visual-language reasoning and open-vocabulary detection to robustly localize $o_k^\star$, effectively disambiguating it from semantic distractors in cluttered scenes.

Equation~\eqref{eq:semantic_grounding} determines \textbf{where} the Task Manager-specified entity resides in the routed visual space. Within each manipulation pattern, the semantic commitment $(\ell_t,o_t^\star)$ is maintained as $(\ell_k,o_k^\star)$ for $t_k \leq t < t_{k+1}$, until the next pattern transition at $t_{k+1}$. The corresponding bounding boxes $\mathbf{b}_{t_k}^v$ provide the initial target localization for the subsequent visual tracking.

\paragraph{\textbf{Goal-Consistent Visual Tracking}}
Once the semantic commitment $(\ell_k,o_k^\star)$ is established, visual tracking maintains its spatial realization as the target location changes with robot motion, object displacement, and occlusion. Rather than repeatedly invoking semantic reasoning, the tracker propagates the grounded target over time, while low-confidence predictions trigger re-grounding to preserve alignment with the committed goal.

Specifically, we initialize the visual tracker SAM2~\cite{ravi2025sam} using the grounding result $\mathbf{b}_{t_k}^{v}$. During execution, each tracker updates the target state as
\begin{equation}
\left(
\mathbf{b}_{t}^{v},
q_{t}^{v}
\right)
=
\mathcal{F}_{v}
\left(
I_{t}^{v},
\mathbf{b}_{t-1}^{v}
\right),
\qquad
v\in\mathcal{V}_k,
\label{eq:visual_tracking}
\end{equation}
where $q_t^v$ denotes the tracking confidence. When $q_t^v$ falls below a predefined threshold, the Semantic Grounder is re-invoked with the same committed target $o_k^\star$ to re-establish its location and reinitialize the tracker. This enables each routed view to continuously follow the intended entity while correcting tracking drift without altering its semantic identity.

\subsection{How: View-Routed Semantic Intent Injection}
We introduce View-Routed Semantic Intent Injection, a visual interface for communicating task-level semantic intent to the action policy without adding trainable parameters to the foundation action model. The key principle is to decouple semantic reasoning from action generation: the high-level perception module determines what is behaviorally relevant and where it appears, while the action policy retains responsibility for deciding how to act. 

Within a manipulation pattern $t_k\leq t<t_{k+1}$, the routed views $\mathcal{V}_k$, semantic commitment $(\ell_k,o_k^\star)$, and tracked states $\{\mathbf{b}_t^v,q_t^v\}_{v\in\mathcal{V}_k}$ define a view-conditioned spatial prior. We realize this principle through two complementary interfaces: pixel-level \textbf{Visual Semantic Highlighting} and token-level \textbf{Semantic Attention Bias}.

\paragraph{\textbf{Pixel-level Visual Semantic Highlighting}}
Although Eq.~\eqref{eq:visual_tracking} represents the tracked state by $\mathbf{b}_t^v$, SAM2 internally produces a dense mask $\mathbf{S}_t^v\in\{0,1\}^{H_v\times W_v}$ for the committed target $o_k^\star$, whose tight enclosing box is $\mathbf{b}_t^v$. We retain this object geometry and highlight the mask only in routed views:
\begin{equation}
\widetilde{I}_t^v
=
\begin{cases}
\left(1-\lambda\mathbf{S}_t^v\right)\odot I_t^v
+\lambda\mathbf{S}_t^v\odot\mathbf{c}, & v\in\mathcal{V}_k,\\
I_t^v, & v\notin\mathcal{V}_k,
\end{cases}
\label{eq:highlight}
\end{equation}
where $\mathbf{c}$ is the rendering color and $\lambda=0.32$ the opacity; a thin contour delineates the target boundary. This translucent cue increases salience without obscuring target appearance or scene context. For the pixel-level interface alone, the policy operates on $\widetilde{\mathcal{I}}_t=\{\widetilde{I}_t^{\mathrm{g}},\widetilde{I}_t^{\mathrm{l}},\widetilde{I}_t^{\mathrm{r}}\}$:
\begin{equation}
\mathbf{a}_t
=\pi_{\theta}\!\left(
\widetilde{\mathcal{I}}_t,\mathbf{s}_t^p,\ell_k
\right).
\label{eq:highlight_policy}
\end{equation}
Beyond indicating \emph{where} to act, the shared overlay partially normalizes target appearance cues such as color into a consistent visual signature, providing the action policy with a target-centric feature that is easier to learn while preserving the underlying object appearance. Because this representation is realized entirely through image-space modification, semantic highlighting remains architecture-agnostic and requires no access to the policy internals.

\paragraph{\textbf{View-Routed Token-Level Attention Prior}}
For the attention-based interface, we keep the input images unchanged and
convert the tracked target mask into a patch-level prior aligned with the
policy's visual tokens. Let
$\overline{\mathbf{S}}_t^v=\mathcal{R}_v(\mathbf{S}_t^v)
\in[0,1]^{H_\pi\times W_\pi}$
denote the mask in view $v$ resized to the policy input resolution. For visual
patch $j$ with image-space support $\Omega_{v,j}$, its target relevance is

\begin{equation}
w_{t,j}^v
=
\begin{cases}
\dfrac{1}{|\Omega_{v,j}|}
\displaystyle\sum_{(x,y)\in\Omega_{v,j}}
\overline{\mathbf{S}}_t^v(x,y),
& v\in\mathcal{V}_k,\ q_t^v\geq\tau_{\mathrm{trk}},\\[6pt]
0, & \text{otherwise},
\end{cases}
\label{eq:patch_weight}
\end{equation}
where $\tau_{\mathrm{trk}}$ is the tracking-confidence threshold used for
semantic re-grounding.

The resulting prior
$\mathbf{w}_t^v=[w_{t,1}^v,\ldots,w_{t,N_v}^v]^\top$
preserves target geometry through fractional patch coverage while assigning
zero relevance to non-routed views. With a $224\times224$ input and
$14\times14$-pixel patch size, each view yields a $16\times16$ grid
($N_v=256$). We inject this spatial prior at \textbf{two complementary levels}: visual self-attention
to shape target-aware representations, and action-to-vision attention to
direct action queries toward target-relevant evidence.

\textbf{Visual Encoder Injection}. Let
$\mathcal{L}_{\mathrm{vis}}\subseteq\{1,\ldots,D_{\mathrm{vis}}\}$
denote the selected injection layers, with
\begin{equation}
g_r^{\mathrm{vis}}
=
\mathbf{1}[r\in\mathcal{L}_{\mathrm{vis}}].
\end{equation}
Rather than treating Transformer depth uniformly, we concentrate injection in
the intermediate layers and retain only sparse injections elsewhere.
This reflects the hierarchical role of depth: early layers preserve local
visual evidence, intermediate layers increasingly integrate task-relevant
semantics, while excessive intervention across all layers can unnecessarily
disturb the pretrained representation hierarchy
\cite{lepori2024beyond,yoo2023improving,xie2026s2vla}.

For visual layer $r$, attention head $h$, query patch $i$, and key patch $j$,
the original pre-softmax attention logit is
\begin{equation}
Z_{r,h}^{\mathrm{vis},v}(i,j)
=
\frac{
\mathbf{Q}_{r,h}^{\mathrm{vis},v}(i)
\mathbf{K}_{r,h}^{\mathrm{vis},v}(j)^\top
}{
\sqrt{d_h^{\mathrm{vis}}}
}
+
\mathcal{M}_{r}^{\mathrm{vis},v}(i,j).
\label{eq:visual_attention_logit}
\end{equation}
We add the spatial prior along the key dimension:
\begin{equation}
\begin{aligned}
B_r^{\mathrm{vis},v}(i,j)
&=
g_r^{\mathrm{vis}} w_{t,j}^v,\\
\widetilde{A}_{r,h}^{\mathrm{vis},v}(i,:)
&=
\operatorname{softmax}
\left(
Z_{r,h}^{\mathrm{vis},v}(i,:)
+
B_r^{\mathrm{vis},v}(i,:)
\right).
\end{aligned}
\label{eq:visual_attention_injection}
\end{equation}
The head-shared bias encourages the model to focus more on target-related patches while retaining information from the surrounding
scene.

\textbf{Action Decoder Injection}. The view-specific priors are aligned with the
policy transformer's complete key sequence:
\begin{equation}
\mathbf{w}_t
=
\operatorname{Concat}
\left(
\mathbf{w}_t^{v^g},
\mathbf{w}_t^{v^l},
\mathbf{w}_t^{v^r},
\mathbf{0}_{N_{\mathrm{nv}}}
\right)
\in[0,1]^{N_K},
\label{eq:token_map}
\end{equation}
where non-visual positions and visual tokens from non-routed views receive zero
bias.

For policy-transformer layer $m$ and attention head $h$, let
\begin{equation}
Z_{m,h}(i,j)
=
\frac{
\mathbf{Q}_{m,h}(i)
\mathbf{K}_{m,h}(j)^\top
}{
\sqrt{d_h}
}
+
\mathcal{M}_m(i,j)
\label{eq:attention_logit}
\end{equation}
denote the original pre-softmax attention logit. Let
$\mathcal{Q}_{\mathrm{act}}$ denote action-query positions and
$\mathcal{L}_{\mathrm{inj}}$ the selected injection layers, following the same
middle-concentrated, sparse-outside schedule. With
$g_m=\mathbf{1}[m\in\mathcal{L}_{\mathrm{inj}}]$, we define
\begin{equation}
\begin{aligned}
B_m(i,j)
&=
\begin{cases}
g_m w_{t,j}, & i\in\mathcal{Q}_{\mathrm{act}},\\
0, & \text{otherwise},
\end{cases}\\[3pt]
\widetilde{A}_{m,h}(i,:)
&=
\operatorname{softmax}
\left(
Z_{m,h}(i,:)+B_m(i,:)
\right).
\end{aligned}
\label{eq:semantic_attention_bias}
\end{equation}

Because $\mathbf{w}_t$ is nonzero only on target-relevant visual keys, the
intervention selectively strengthens action-to-target attention while
preserving the policy's original visual-language interactions. Combining both intent-injection interfaces, the complete HINT
policy is
\begin{equation}
\mathbf{a}_t
=
\pi_\theta\!\left(
\widetilde{\mathcal{I}}_t,\mathbf{s}_t^p,\ell_k;
\{B_r^{\mathrm{vis},v}\}_{r,v},
\{B_m\}_m
\right).
\label{eq:hint_policy}
\end{equation}

%% file: 4_experiment.tex
\section{Experiments}

\begin{figure*}[t!]
    \centering
    \vspace{-1.5em}
    \includegraphics[width=1.0\linewidth]{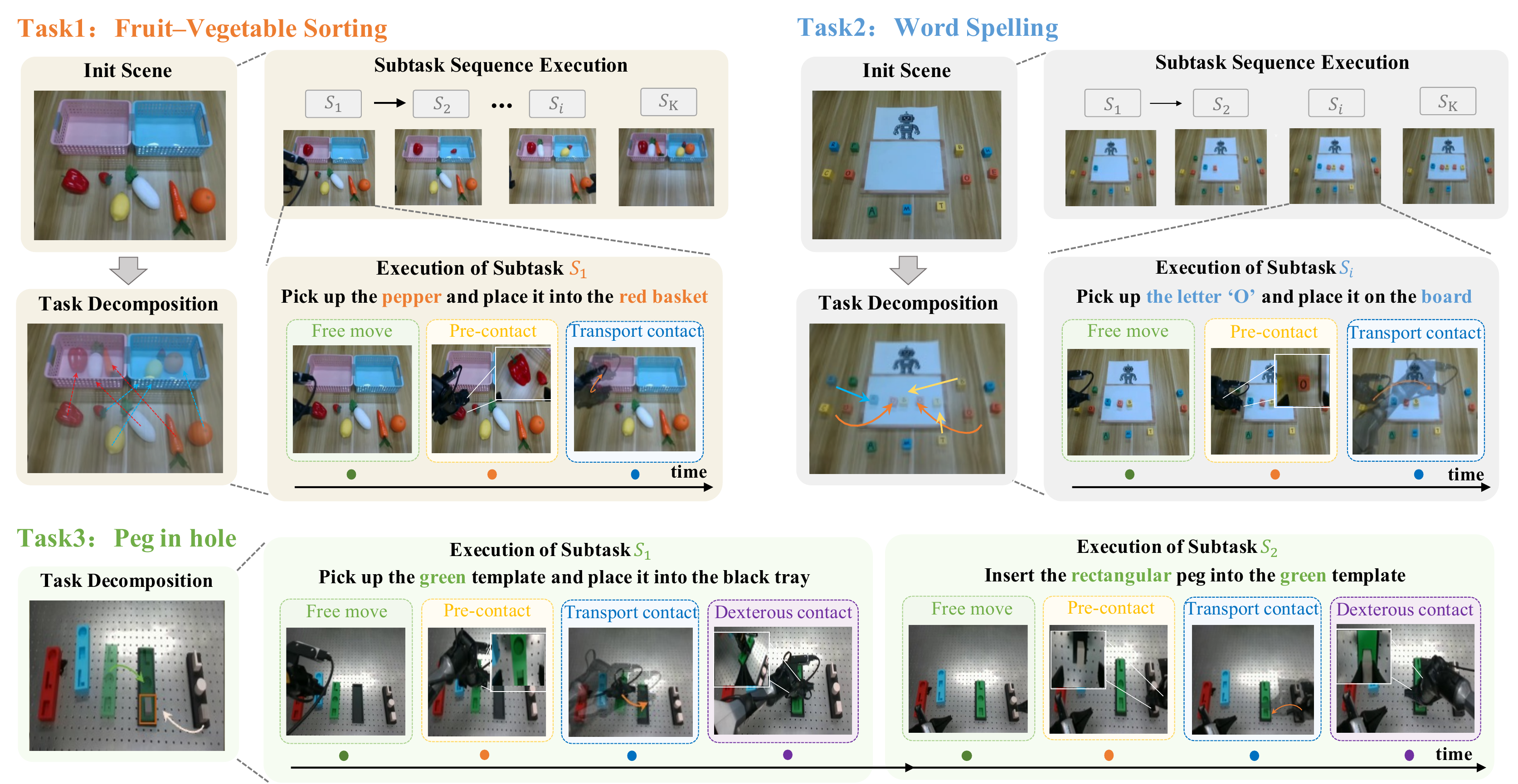}
    \caption{Evaluation tasks and their decomposition into manipulation
    patterns. Fruit-vegetable sorting (top left) and word spelling (top right)
    require a sequence of pick-and-place subtasks, each comprising free move,
    pre-contact, and transport contact. Peg-in-hole insertion (bottom) further
    includes dexterous contact for contact-rich placement and insertion. $S_i$
    denotes the $i$th semantic subtask; colored outlines and timelines indicate
    the manipulation pattern active during each execution segment.}
    \label{fig:tasks}
\end{figure*}

We organize the evaluation around four questions while retaining a conventional
setup-results-ablation structure: \textbf{(Q1)} Can \textbf{HINT} effectively convey task semantics while preserving accurate action control?
\textbf{(Q2)} Does \textbf{HINT} improve
semantic-compositional generalization under out-of-distribution conditions while the required motor primitives remain familiar? \textbf{(Q3)} Can a shared
manipulation-pattern representation support subtask switching and reliable online
routing across the evaluated tasks? \textbf{(Q4)} Does pattern-triggered multi-view reasoning improve efficiency
without sacrificing task performance?

\begin{table*}[t]
\centering
\caption{Performance on the three in-distribution long-horizon manipulation
tasks. IS is the percentage of correct semantic decisions and target selections;
Sub.\ SR is the percentage of required subtasks completed; and Full SR is the
percentage of trials in which the complete task is successful.}
\label{tab:main_results}
\resizebox{\textwidth}{!}{
\begin{tabular}{lccccccccc}
\toprule
\multirow{2}{*}{\textbf{Method}}
& \multicolumn{3}{c}{\textbf{Fruit-Vegetable Sorting}}
& \multicolumn{3}{c}{\textbf{Word Spelling}}
& \multicolumn{3}{c}{\textbf{Peg-in-Hole Insertion}} \\
\cmidrule(lr){2-4}\cmidrule(lr){5-7}\cmidrule(lr){8-10}
& IS $\uparrow$ & Sub.\ SR $\uparrow$ & Full SR $\uparrow$
& IS $\uparrow$ & Sub.\ SR $\uparrow$ & Full SR $\uparrow$
& IS $\uparrow$ & Sub.\ SR $\uparrow$ & Full SR $\uparrow$ \\
\midrule
Wall-OSS-0.5~\cite{yu2026walloss05technicalreport}
& 53.1\% & 53.1\% & 0.0\%
& 44.7\% & 36.2\% & 13.3\%
& 65.0\% & 25.0\% & 5.0\% \\
$\pi_{0.5}$~\cite{black2025pi05}
& 52.2\% & 52.2\% & 10.0\%
& 44.7\% & 42.6\% & 13.3\%
& 35.0\% & 27.5\% & 5.0\% \\
\midrule
Wall-OSS-0.5 + HINT
& 85.8\% & 85.8\% & 40.0\%
& 83.0\% & 80.9\% & 40.0\%
& 86.7 \% & 47.5 \% & 10.0\% \\
$\pi_{0.5}$ + HINT
& \textbf{91.2\%} & \textbf{91.2\%} & \textbf{60.0\%}
& \textbf{97.9\%} & \textbf{95.7\%} & \textbf{86.7\%}
& \textbf{100.0\%} & \textbf{67.5\%} & \textbf{40.0\%} \\
\bottomrule
\end{tabular}}
\end{table*}

\subsection{Setup}

\textbf{Platform.}
The robotic platform consists of two AgileX PiPER arms configured as a dual-arm
manipulation system. Three Intel RealSense D435 RGB-D cameras provide one global
view of the workspace and two wrist-mounted views attached to the left and right
arms, respectively.

\textbf{Tasks.}
As shown in Fig.~\ref{fig:tasks}, we evaluate three language-conditioned tasks:
\textit{fruit-vegetable sorting}, \textit{word spelling}, and
\textit{peg-in-hole insertion}. Each task requires the robot to interpret a
high-level instruction, resolve a sequence of task-dependent semantic targets,
and execute multiple manipulation patterns. The first two tasks emphasize
repeated semantic target changes and object transport, whereas peg-in-hole
insertion additionally includes the \textit{dexterous contact} pattern and thus
tests contact-rich execution after semantic target selection.

\textbf{Data Collection and Annotation.}
We collect demonstrations through arm-to-arm teleoperation at 30~Hz, recording multi-view images, robot actions, proprioceptive states, and joint
torques. According to task complexity and precision requirements, we collect 50
demonstrations for fruit-vegetable sorting, 80 for word spelling, and 150 for
peg-in-hole insertion.

\textbf{Baselines.}
We evaluate HINT on two large-scale action-pretrained foundation policies, $\pi_{0.5}$~\cite{black2025pi05} and Wall-OSS-0.5~\cite{yu2026walloss05technicalreport}, to assess HINT as a plug-and-play semantic interface for strong pretrained action models. For each backbone, the base and HINT variants use the same demonstrations, optimization protocol, and action-backbone initialization, while HINT introduces no additional trainable parameters into the foundation action backbone. This backbone-matched setup isolates the contribution of the HINT from differences in training data or underlying action capability.

Representative long-horizon and grounding methods, such as OneTwoVLA~\cite{lin2025onetwovla} and VP-VLA~\cite{wang2026vpvla}, differ substantially in action-model pretraining, architecture, and training paradigm, making direct quantitative comparisons less controlled for this purpose. Our primary quantitative evaluation mainly addresses whether HINT can consistently improve semantic alignment and long-horizon execution when applied to the same large-scale pretrained action backbone.

\textbf{Metrics.}
Building on VLABench~\cite{vlabenchmark}, we report Intention Score (IS), Subtask
Success Rate (Sub.\ SR), and Full-Task Success Rate (Full SR). IS is the
proportion of correct semantic decisions
$\mathrm{IS}=n_{\mathrm{correct}}/N$, whereas Sub.\ SR is the fraction of
required subtasks completed successfully,
$\mathrm{Sub.\,SR}=m_{\mathrm{done}}/M$. Full SR is the percentage of trials
in which the entire task is completed. Because a single local failure can reduce
Full SR, particularly in long-horizon and contact-rich tasks, we interpret it
together with the IS and Sub.\ SR metrics.

\textbf{Evaluation Protocol.}
All experiments are conducted on a workstation with two NVIDIA RTX 3090 GPUs
under identical workspace configurations and predefined randomized test
settings. We conduct 20 trials for fruit-vegetable sorting and peg-in-hole
insertion and 15 trials for word spelling. Each out-of-distribution (OOD) variant
and ablation is evaluated over 10 trials. Full SR is a deliberately strict
end-to-end metric: a trial is counted as successful only if every required
subtask is completed without failure. For word spelling and peg-in-hole insertion, we assign task-specific partial credit of 0.5; exact criteria are given in the supplement.

\subsection{Results}

\begin{table*}[t]
\centering
\caption{Semantic-compositional OOD evaluation on three manipulation tasks. Each method is evaluated over 10 trials per task; the required motor primitives remain familiar while objects, attributes, layouts, goals, or instructions change.}
\label{tab:unseen_generalization}
\resizebox{\textwidth}{!}{
\begin{tabular}{lccccccccc}
\toprule
\multirow{2}{*}{\textbf{Method}}
& \multicolumn{3}{c}{\textbf{Fruit-Vegetable Sorting OOD}}
& \multicolumn{3}{c}{\textbf{Word Spelling OOD}}
& \multicolumn{3}{c}{\textbf{Peg-in-Hole Insertion OOD}} \\
\cmidrule(lr){2-4}\cmidrule(lr){5-7}\cmidrule(lr){8-10}
& IS $\uparrow$ & Sub.\ SR $\uparrow$ & Full SR $\uparrow$
& IS $\uparrow$ & Sub.\ SR $\uparrow$ & Full SR $\uparrow$
& IS $\uparrow$ & Sub.\ SR $\uparrow$ & Full SR $\uparrow$ \\
\midrule
Wall-OSS-0.5~\cite{yu2026walloss05technicalreport}
& 47.8\% & 37.0\% & 0.0\%
& 36.4\% & 31.8\% & 0.0\%
& 50.0\% & 25.0\% & 10.0\% \\
$\pi_{0.5}$~\cite{black2025pi05}
& 56.5\% & 52.2\% & 0.0\%
& 40.9\% & 36.4\% & 0.0\%
& 26.7\% & 20.0\% & 0.0\% \\
\midrule
Wall-OSS-0.5 + HINT
& 73.9\% & 65.2\% & 20.0\%
& 81.8 \% & 72.7\% & 20.0\%
& 83.3 \% & 30.0 \% & 10.0\% \\
$\pi_{0.5}$ + HINT
& \textbf{82.6\%} & \textbf{80.4\%} & \textbf{30.0\%}
& \textbf{95.5\%} & \textbf{87.5\%} & \textbf{30.0\%}
& \textbf{90.0\%} & \textbf{55.0\%} & \textbf{30.0\%} \\
\bottomrule
\end{tabular}}
\end{table*}

\textbf{HINT turns high-level instruction into executable intent (Q1).} Table~\ref{tab:main_results} reveals a fundamental mismatch between the semantic capacity of foundation policies and their operationalization in long-horizon manipulation. Whether given high-level instructions or fine-grained subtask prompts, baselines struggle to determine \emph{what} to manipulate. For example, in word spelling, they often rely on incidental visual correlations like block color rather than instructed letter identities. Similarly, target selection in peg-in-hole insertion remains inconsistent. Pretrained VLMs provide broad semantic priors, but adapting their representations to robot dynamics can weaken the structure that supports semantic generalization. This indicates an interface bottleneck rather than a lack of manipulation primitives: \textbf{preserving both pretrained semantics and effective control within a single policy is difficult.}

HINT addresses this bottleneck by making semantic target information spatially explicit to the action policy without introducing additional trainable parameters into the foundation action model. It improves the Intention Score (IS), raising it for Wall-OSS-0.5 and $\pi_{0.5}$ by up to 34.9 and 45.9 percentage points in sorting and spelling, and boosting $\pi_{0.5}$'s IS from 35.0\% to 100.0\% in peg-in-hole insertion. The consistent gains across policies and tasks indicate that the effect is not specific to a single action backbone. Instead, continuous semantic intent injection provides an interface between semantic target grounding and low-level motor execution.

The close alignment between IS and Subtask Success Rate (Sub.\ SR) supports this. With HINT, $\pi_{0.5}$ achieves Sub.\ SRs of 91.2\% and 95.7\% in sorting and spelling, nearly matching their IS values (91.2\% and 97.9\%). This correspondence indicates that once the correct semantic target is established, the policy can usually execute it without degrading its motor competence. Peg-in-hole insertion presents a different regime: HINT raises IS to 100.0\%, while Sub.\ SR and Full SR reach 67.5\% and 40.0\%, respectively. The remaining gap is primarily action-side: insertion requires higher geometric precision and contact correction than the base policy can reliably provide.

\begin{figure}[t!]
    \centering
    \includegraphics[width=1.0\linewidth]{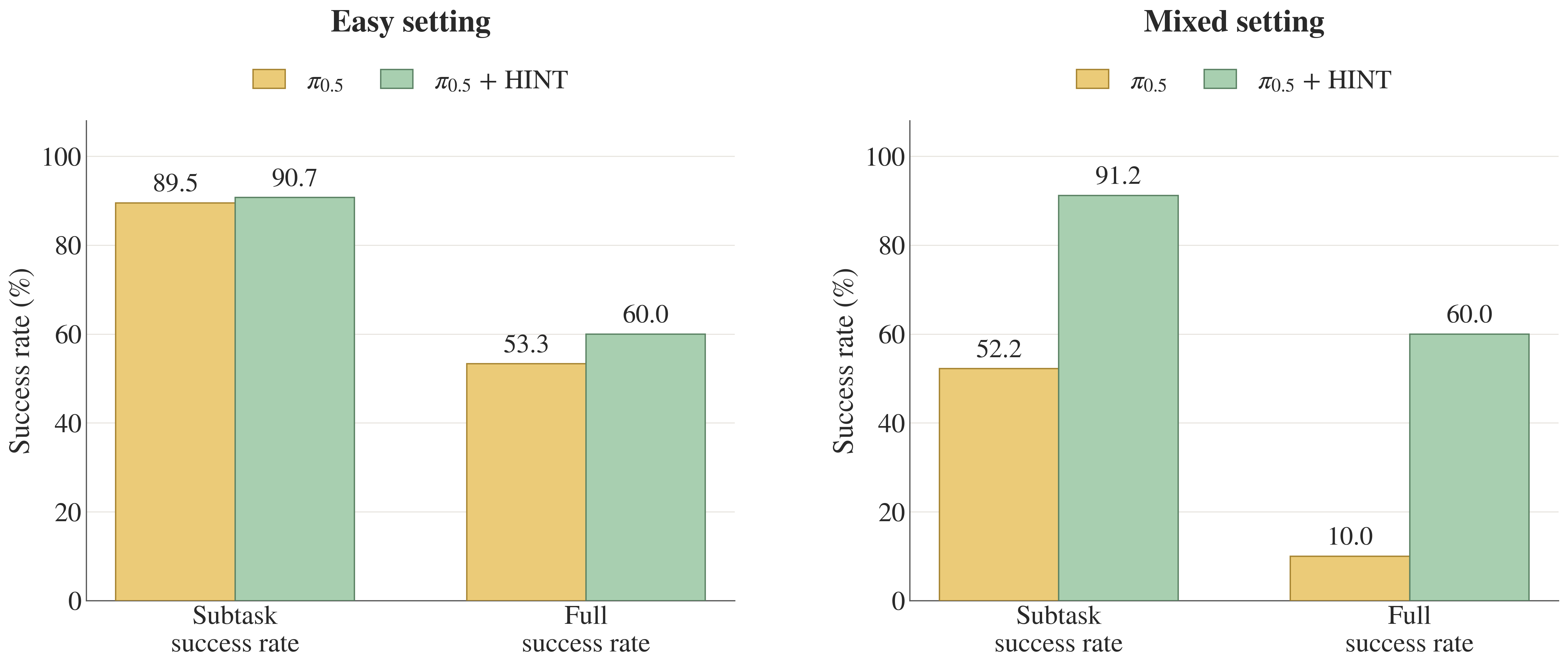}
    \caption{Effect of visual shortcuts on fruit-vegetable sorting. In the
    easy setting (left), fruits always map to the blue basket and vegetables to
    the red basket; in the mixed setting (right), the instruction specifies a
    variable category-to-basket mapping.}
    \label{fig:vision_shortcut}
    \vspace{-1.5em}
\end{figure}

\textbf{The easy-mixed contrast exposes visual shortcuts in VLA policies (Q1).}
To distinguish language-conditioning failure from insufficient manipulation capability, we compare two sorting settings with identical scenes. The easy setting maintains a fixed category-to-color mapping while randomizing container positions, whereas the mixed setting changes this mapping according to the instruction. As shown in Fig.~\ref{fig:vision_shortcut}, the base $\pi_{0.5}$ achieves 89.5\% Sub.\ SR in the easy setting, confirming that the pretrained policy already possesses strong visual matching and precise action-generation capabilities. However, its performance drops to 52.2\% in the mixed setting, close to binary guessing. The policy therefore knows \emph{where} and \emph{how} to act, but fails to use language reliably to determine \emph{which} target is intended. \textbf{This reveals a visual shortcut: when vision and language conflict, the visually learned correspondence dominates action generation}, consistent with the language-conditioning degradation observed in prior VLA studies~\cite{yuan2026qwenrobotmanip}.

HINT performs similarly to the base policy in the "Easy" setting but improves Sub.\ SR by 38.5 percentage points in the "Mixed" setting. This selectivity is more informative than the aggregate gain: HINT adds little when vision alone determines the action, yet restores performance when intent must override a misleading visual association. Unlike approaches that address language degradation through large-scale pretraining or vision-language co-training~\cite{yuan2026qwenrobotmanip}, \textbf{HINT introduces an agentic intent injection framework on top of standard action-model training}. By expressing intent through the visual channel already used by the action policy, it improves intent-action alignment without adding trainable modules to the foundation action model.

\begin{figure}[t!]
    \centering
    \includegraphics[width=1.0\linewidth]{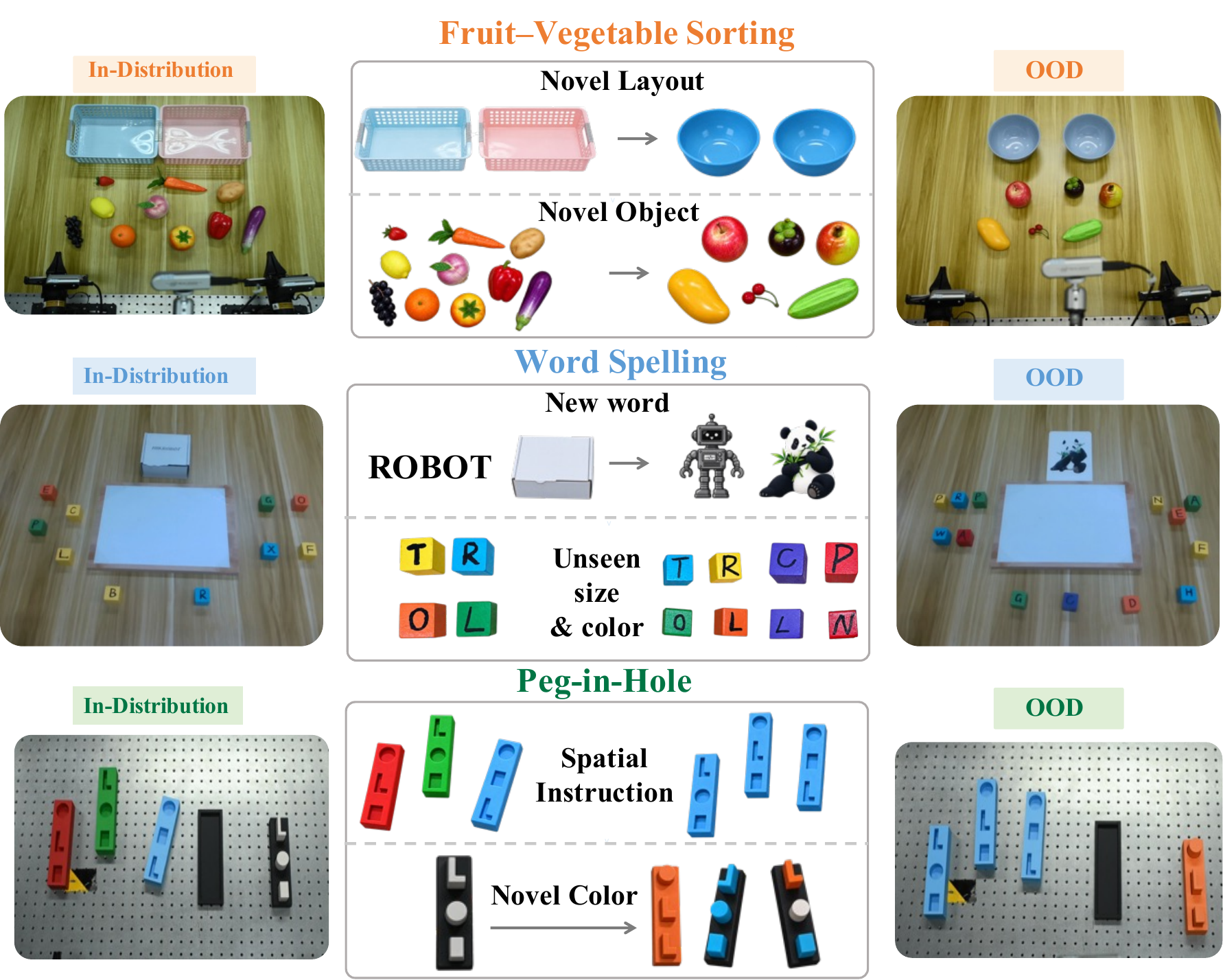}
    \caption{Three task out-of-distribution (OOD) test
    configurations. We evaluate generalization across unseen objects, goal representations, visual attributes, spatial layouts, and semantic instructions.}
    \label{fig:OOD_experiment}
    \vspace{-1.5em}
\end{figure}

\textbf{HINT aligns familiar manipulation primitives with unseen semantic compositions (Q2).} We evaluate compositional out-of-distribution tasks in which the semantic configuration changes while the required manipulation primitives remain familiar. As summarized in Fig.~\ref{fig:OOD_experiment}, word spelling introduces unseen letters, colors, and layouts; sorting replaces the training categories and containers with novel objects and target plates; and insertion requires unseen peg-base combinations or spatial relations such as left, right, above, and below. These shifts alter \emph{what} the policy should act on, allowing us to test whether HINT can bind familiar actions to previously unseen intent without requiring new low-level manipulation capabilities.

The corresponding results in Table~\ref{tab:unseen_generalization} show that, across the three OOD tasks, HINT raises the mean IS of $\pi_{0.5}$ from 41.4\% to 89.4\% and the mean Sub.\ SR from 36.2\% to 74.3\%, while achieving a 30.0\% Full SR on each task compared with 0.0\% for the base $\pi_{0.5}$. The improvements on Wall-OSS-0.5 further indicate that the semantic benefit is not specific to one action backbone. \textbf{HINT better leverages this pretrained semantic generalization} and exposes it to the action policy, binding novel categories, attributes, goal objects, and relations to familiar manipulation primitives. Full-task success requires all subtasks to succeed in the correct order (e.g., the full letter sequence in spelling). Peg insertion is further limited by the precision required for contact-rich control: HINT improves instruction-conditioned object grounding, but does not target fine-grained contact execution.

\begin{figure}[t!]
    \centering
    \includegraphics[width=1.0\linewidth]{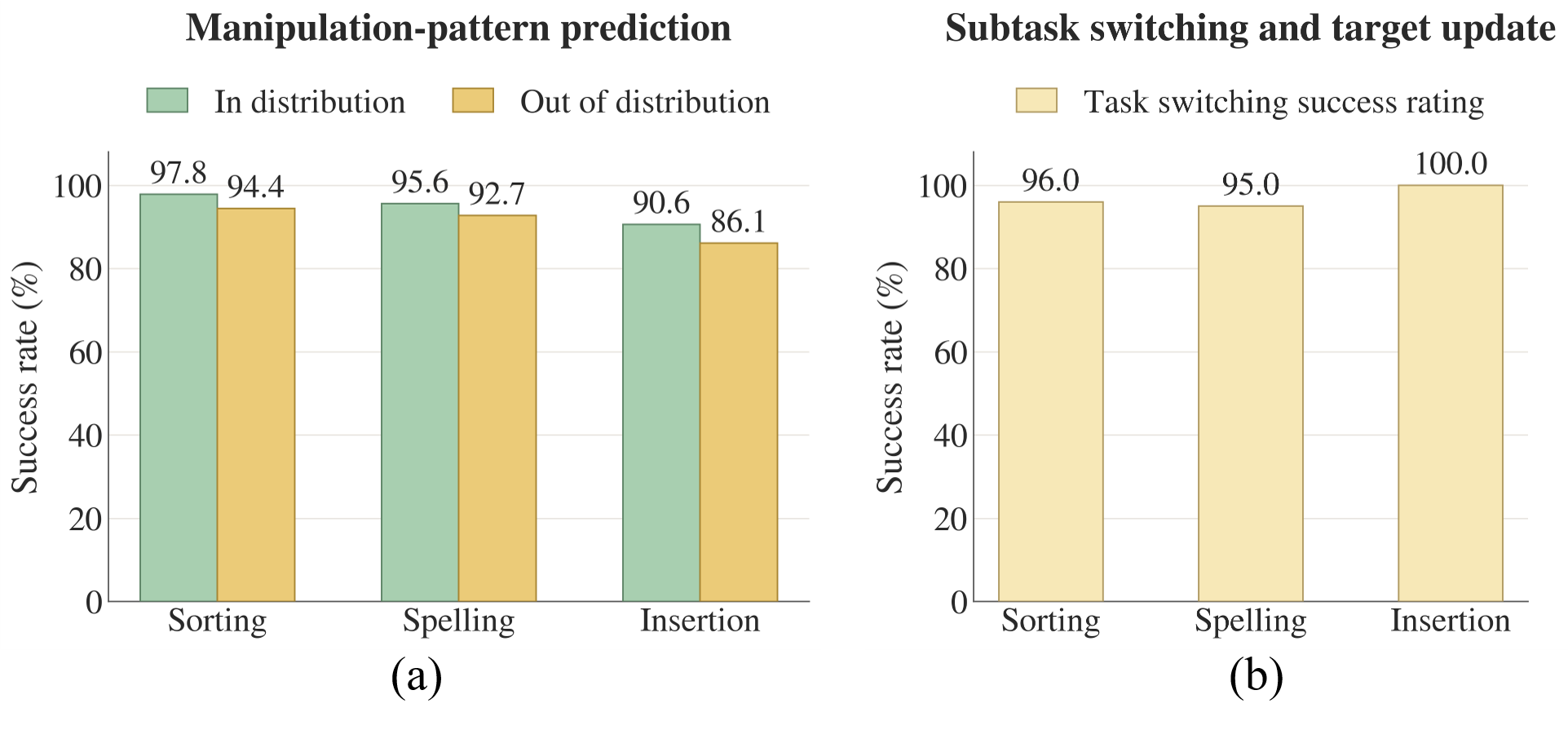}
    \vspace{-1.5em}
    \caption{Reliability of manipulation-pattern-based online routing.
    (a) Manipulation-pattern prediction success rate under ID and OOD
    conditions, evaluated over 180 trials for each task-condition pair.
    (b) Subtask-switch success rate aggregated over ID and OOD conditions and
    evaluated over 100, 100, and 60 transitions for sorting, spelling, and
    insertion, respectively. A switch is successful only if the transition is
    detected and the correct next semantic target is activated.}
    \label{fig:pattern_SR}
\end{figure}

\textbf{Manipulation patterns drive online routing (Q3).} We evaluate the manipulation-pattern router of action execution using a single Pattern Router jointly trained on demonstrations from all three tasks. Figure~\ref{fig:pattern_SR} summarizes both pattern prediction and subtask-switch reliability. In Figure~\ref{fig:pattern_SR}(a), the mean Pattern Success Rate decreases by only 3.6 percentage points from ID to OOD conditions (94.7\% to 91.1\%), despite substantial changes in object identity, appearance, and semantic relations, indicating that the shared representation primarily captures interaction phase rather than task-specific visual content. Figure~\ref{fig:pattern_SR}(b) further shows 95-100\% Switch SR, with 251 of 260 transitions correctly activating the next semantic target. Peg-in-hole insertion yields lower Pattern SR (90.6\% ID and 86.1\% OOD) due to unstable, force-sensitive contact near phase boundaries, yet still achieves 100\% switching, suggesting that local pattern ambiguity does not propagate into semantic transition errors. Together, these results establish manipulation patterns as a reliable event interface between continuous execution and discrete reasoning, determining \emph{which view} to inspect, \emph{when} to update semantic grounding or switch subtasks, while leaving \emph{how} the selected action is executed to the action policy.

\textbf{Pattern-triggered reasoning aligns computation with semantic change (Q4).}
Table~\ref{tab:inference_latency} decomposes the latency of semantic reasoning
and continuous tracking. Grounding DINO \cite{liu2024groundingdino} handles direct object grounding,
whereas Qwen3 resolves fine-grained attributes, spatial relations, and
compositional semantics. Qwen reasoning frame requires 630 ms, 3.4$\times$ the latency of the base
policy, reflecting the capability-latency trade-off also observed in
multi-expert VLAs such as UAM~\cite{zhang2026uam}. HINT avoids placing this cost
on every frame. The 14-ms pattern router invokes semantic reasoning only at
pattern transitions, while SAM2 maintains the selected target between them.
Consequently, tracking reduces frame latency by 61.6\% relative to Qwen
reasoning and, after excluding policy inference, reduces auxiliary overhead
from 445 ms to 57 ms. HINT therefore preserves full semantic reasoning while
making its invocation sparse. The low-latency schedule is also compatible with asynchronous executors such as RTC and DTW~\cite{rtc,forcepolicy}. Pattern-triggered multi-view reasoning reduces repeated semantic inference, supporting smoother closed-loop execution with less trajectory interruption.

\begin{table}[t]
\centering
\caption{Inference time for each module}
\label{tab:inference_latency}
\resizebox{0.85\columnwidth}{!}{
\begin{tabular}{lc}
\toprule
\textbf{Module / Frame} & \textbf{Inference time} $\downarrow$ \\
\midrule
Base Policy ($\pi_{0.5}$)              & 185 ms \\
Pattern Router                         & 14 ms  \\
Visual Tracker (SAM2)                  & 43 ms  \\
Object-level Grounder (Grounding DINO) & 227 ms \\
Fine-grained Semantic Grounder (Qwen)  & 431 ms \\
\midrule
Reasoning Frame (DINO / Qwen)           & 426 / 630 ms \\
Tracking Frame                          & 242 ms \\
\bottomrule
\end{tabular}}
\vspace{-1em}
\end{table}

\subsection{Ablation Study}
\label{sec:ablation}

\textbf{Dual-path semantic injection improves deployment robustness (Q1,Q2).}
We ablate the two injection pathways on OOD word spelling. The attention-only
variant removes input highlighting, whereas the highlighting-only variant
removes the internal attention prior. All variants use the same
$\pi_{0.5}$ training protocol, add no trainable parameters for semantic
injection, and are evaluated over 10 trials.

The single-path variants expose complementary failure modes. Highlighting
suppresses appearance-specific variation and provides unseen targets with a
stable external referent, strengthening semantic commitment but occasionally
resulting in incorrect picks, misplaced blocks, or overturned blocks.
Attention injection biases the action expert toward target-relevant patches,
allowing the attention computation to associate the object with the
corresponding action. However, because this guidance remains implicit, it is
less reliable when multiple visually similar OOD letters are present, often resulting in missed grasps. HINT therefore combines an explicit, invariant representation of
\emph{what} to manipulate with an internal focus on \emph{how} that target
relates to action, jointly preserving long-horizon intent and control fidelity.

As summarized in Table~\ref{tab:semantic_guidance_ablation}, the complete design combines an appearance-invariant external referent with an
internal target-focused prior, increasing IS to 95.5\%, Sub.\ SR to 87.5\%,
and Full SR to 30.0\%, higher than either single-path variant. The ablation
therefore answers Q1 and Q2 jointly: highlighting stabilizes intent under
unseen visual semantics, while attention preserves the spatial evidence needed
for accurate control. Their combination conveys OOD intent without sacrificing
the action policy's control fidelity.

\begin{figure}[t!]
\centering
\includegraphics[width=1.0\linewidth]{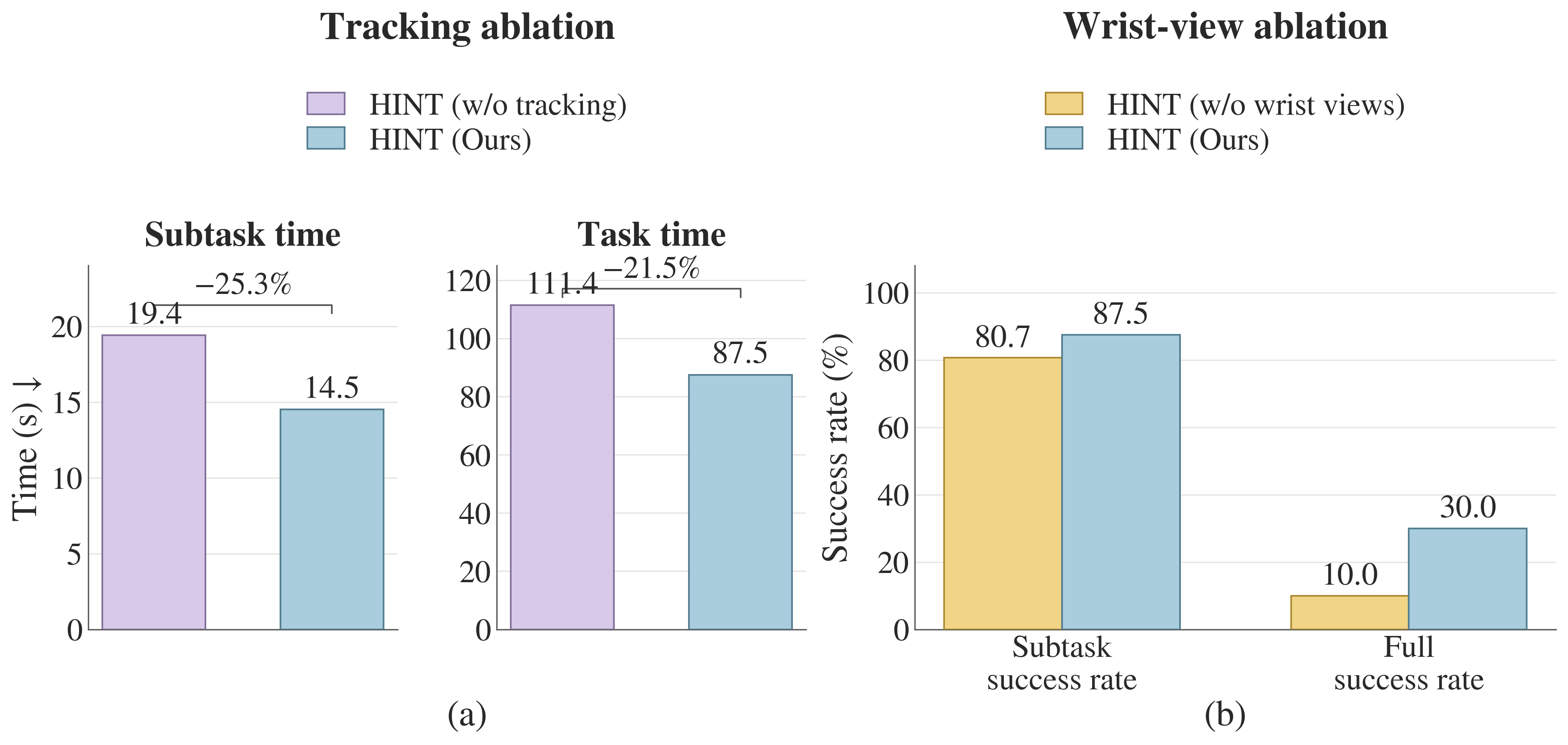}
\caption{Independent ablations of continuous tracking and wrist-view perception on OOD word spelling. Tracking reduces completion time, while wrist views improve subtask and full-task success rates.}
\label{fig:multi_view_ablation}
\end{figure}

\begin{table}[t]
\centering
\caption{Ablation of the two semantic-guidance pathways on OOD word spelling.
All variants use $\pi_{0.5}$ as the base policy.}
\label{tab:semantic_guidance_ablation}
\resizebox{\columnwidth}{!}{
\begin{tabular}{lccc}
\toprule
\textbf{Method} & \textbf{IS} $\uparrow$ & \textbf{Sub.\ SR} $\uparrow$
& \textbf{Full SR} $\uparrow$ \\
\midrule
$\pi_{0.5}$~\cite{black2025pi05}
& 40.9\% & 36.4\% & 0.0\% \\
HINT {\footnotesize (w/o Highlighting)}
& 59.1\% & 58.0\% & 10.0\% \\
HINT {\footnotesize (w/o Attention)}
& 88.6\% & 73.9\% & 10.0\% \\
HINT {\footnotesize (Ours)}
& \textbf{95.5\%} & \textbf{87.5\%} & \textbf{30.0\%} \\
\bottomrule
\end{tabular}}
\vspace{-1em}
\end{table}

\textbf{Tracking and wrist views improve efficiency and reliability (Q4).}
We evaluate OOD word spelling, where dense layouts and changing targets challenge both local visibility and temporal consistency. We separately ablate continuous tracking and wrist-view perception. Without tracking, the target is re-grounded through semantic reasoning every two frames; without wrist views, perception relies only on the global camera. All other components remain unchanged, and the pattern router is used only for subtask switching.

As shown in Fig.~\ref{fig:multi_view_ablation}, continuous tracking reduces subtask and full-task completion times by 25.3\% and 21.5\%, respectively, by preserving target correspondence instead of repeatedly reconstructing it through semantic inference. Separately, wrist views increase subtask success from 80.7\% to 87.5\% and full-task success from 10.0\% to 30.0\%, showing local wrist observations are important for maintaining reliable intent alignment. Together, these results highlight two complementary principles: \textbf{visual persistence improves efficiency, while viewpoint diversity improves robustness.}

%% file: 5_conclusion.tex
\section{Conclusion}

This work studies how to align robot action policies with human intent while preserving control efficiency and low inference latency. In VLA and related models, language provides a sparse conditioning signal, while visual observations continuously drive action generation. As a result, changes in language are often overlooked, and actions follow learned visual shortcuts rather than human intent. HINT converts sparse language instructions into continuous, target-specific visual guidance. At manipulation-pattern transitions, semantic reasoning determines what the robot should act on; between transitions, visual tracking maintains where the target is relative to the robot. We explore two visual intent interfaces that provide this guidance across different foundation action models, keeping human intent aligned with action throughout execution. Experiments across three long-horizon tasks show that existing policies can rely on visual shortcuts, whereas visual intent injection substantially improves target selection, task progress, and success under both in-distribution and semantic-compositional out-of-distribution configurations.

%% file: 6_limitations.tex
\section{Limitations and Future Work}

\textbf{The Boundary of Semantic Generalization.}
The generalization of HINT partly inherits the semantic priors of the underlying VLM, whose coverage may become less reliable under novel concepts or ambiguous observations. Such semantic uncertainty can propagate to subtask inference and target grounding. Future work may explore uncertainty-aware reasoning and online semantic verification to further extend HINT toward open-world manipulation.

\textbf{Scaling the Shared Pattern Router.}
Our current Pattern Router is a single lightweight model trained jointly on demonstrations from all three evaluated tasks, and it exhibits reliable prediction across their ID and semantic OOD conditions. We have not yet evaluated whether the same network capacity remains sufficient when the pattern representation is pretrained over a substantially larger and more heterogeneous task collection. Large-scale pattern pretraining may require richer representations or increased router capacity.

\textbf{Additional Data Preparation.}
HINT requires manual pattern annotations to train the lightweight Pattern Router. In addition, standard action-policy training uses an auxiliary preprocessing pipeline to generate highlighted observations and attention-prior maps. Future work could reduce the reliance on manual pattern labels and simplify this preprocessing pipeline.

%% file: supplementary.tex

\section{Implementation Details}
\label{sec:supp_impl}

\subsection{Semantic-Interface Preparation}
The base-policy and \ours{} variants share demonstrations, optimization, and
action-backbone initialization; only the policy-facing semantic interface
changes. For \ours{}, training observations are rendered offline from
ground-truth pattern and target annotations. Deployment applies the same
rendering and patch-prior construction to masks produced by online grounding
and tracking. This train--test symmetry is essential: the overlay and attention
prior are part of the policy input, not auxiliary supervision, and introduce no
trainable parameters into the foundation action backbone.

\begin{figure*}[!t]
\centering
\includegraphics[width=\textwidth]{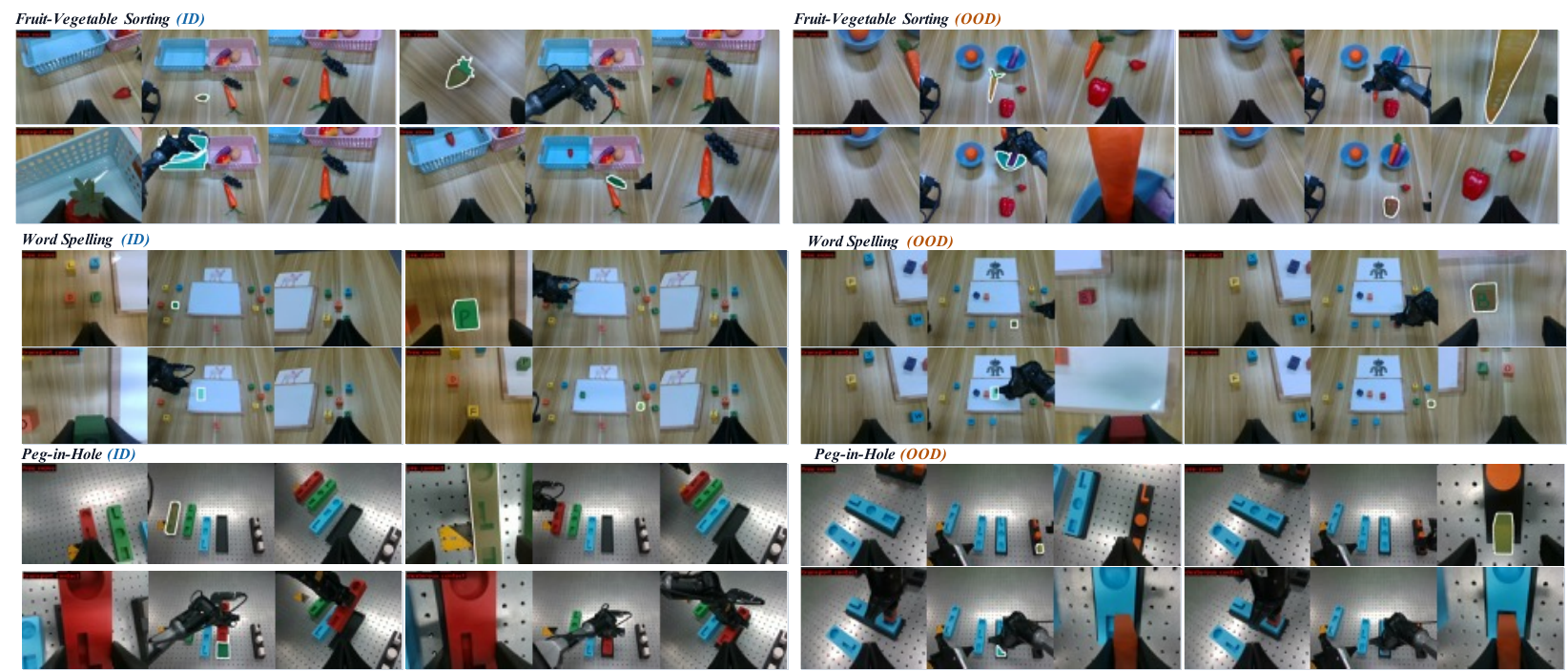}
\caption{Grounding and policy-facing highlighting for sorting, spelling, and
peg-in-hole insertion under ID and OOD variations. Each overlay identifies the
entity selected by the Task Manager in the view routed by the current
manipulation pattern.}
\label{fig:highlighted_inputs}
\end{figure*}

The input resolution is $224\!\times\!224$ for \pis{} and
$448\!\times\!448$ for Wall-OSS-0.5. Highlighting uses RGB $(0,220,120)$ with
opacity $0.32$, a white contour, and a 3-pixel contour width. Fractional target
coverage is computed on a $16\!\times\!16$ patch grid for \pis{} and a
$32\!\times\!32$ grid for Wall-OSS-0.5. Figure~\ref{fig:highlighted_inputs}
shows the resulting policy inputs; Table~\ref{tab:policy_train} records the
backbone-specific fine-tuning settings. Both backbones use
AdamW~\cite{loshchilov2019decoupled} with cosine decay
\cite{loshchilov2017sgdr};

\begin{table}[H]
\centering
\caption{Action-policy fine-tuning hyperparameters.}
\label{tab:policy_train}
\scriptsize
\setlength{\tabcolsep}{2.0pt}
\begin{tabularx}{\columnwidth}{@{}lYY@{}}
\toprule
Hyperparameter & \pis{} & Wall-OSS-0.5 \\
\midrule
Initialization & \texttt{pi05\_base} & Wall-OSS-0.5 \\
Input / patch grid & $224^2$ / $16^2$ & $448^2$ / $32^2$ \\
Optimizer & AdamW & AdamW, $\beta=(.9,.95)$ \\
Weight decay / $\epsilon$ & -- / -- & $10^{-8}$ / $10^{-8}$ \\
Learning rate & $2\!\times\!10^{-5}\!\rightarrow\!2\!\times\!10^{-7}$ & $5\!\times\!10^{-5}\!\rightarrow\!10^{-6}$ \\
Warm-up / schedule & 5k / cosine (50k) & 1k / cosine (200k) \\
Gradient clip & 1.0 & 1.0 \\
EMA & 0.999 & -- \\
Steps (sort / spell / peg) & 50k / 80k / 80k & 50k / 80k / 80k \\
Global batch & 32 & 32 ($4\!\times\!2$ GPUs $\times\!4$ accum.) \\
Sharding / precision & FSDP / -- & FSDP / bfloat16 \\
\bottomrule
\end{tabularx}
\end{table}

\subsection{Pattern Router}
To complement the architectural overview in the main paper, the Router consumes
the current frame from all three cameras together with a 12-frame ($0.4$~s)
history of proprioception and joint effort. RGB inputs are bilinearly resized to
$120\!\times\!160$, antialiased, mapped to $[-1,1]$, and ImageNet-normalized in
the backbone. No stochastic image augmentation is used. Each view has a residual
two-layer convolutional adapter, followed by a shared ImageNet-pretrained
ResNet-18~\cite{deng2009imagenet,he2016deep}. The stem and layers 1--2 are
frozen; layers 3--4 are fine-tuned. Each view is projected to 128 dimensions.
Proprioception and effort are embedded independently to 64 dimensions and
encoded by separate one-layer, 128-D GRUs~\cite{cho2014learning}. Softmax-gated
fusion maps the five streams to a single 128-D representation.

The output comprises six view-resolved pattern logits and four sigmoid progress
heads, one for each pattern family. Data are split by episode rather than by
frame, preventing near-duplicate temporal observations from crossing the
train--validation boundary. Frames are sampled uniformly without
pattern-frequency rebalancing. Table~\ref{tab:router_train} gives the remaining settings.

\begin{table}[H]
\centering
\caption{Pattern Router training hyperparameters.}
\label{tab:router_train}
\footnotesize
\setlength{\tabcolsep}{3.2pt}
\begin{tabularx}{\columnwidth}{@{}lY@{}}
\toprule
Hyperparameter & Value \\
\midrule
Train / validation & 90\% / 10\% episodes \\
Seed & 0 \\
Optimizer / weight decay & AdamW / $5\!\times\!10^{-4}$ \\
Learning rate & $5\!\times\!10^{-5}\!\rightarrow\!3\!\times\!10^{-7}$ (cosine) \\
Warm-up / epochs & 5 / 50 \\
Batch size & 64 \\
Label smoothing / grad. clip & 0.05 / 5.0 \\
Input history & RGB $\times1$; state/effort $\times12$ \\
Loss & weighted CE $+\,0.5\,$weighted MSE \\
\bottomrule
\end{tabularx}
\end{table}

At inference, a proposed transition into or out of \emph{free move} is accepted
when the current-pattern progress exceeds 0.7 and the candidate-pattern progress
is below 0.3. A candidate that persists for five frames overrides this gate,
preventing local boundary noise from indefinitely delaying a valid event.
Transitions among contact-side patterns are not progress-gated. This temporal
filter affects only perception scheduling and semantic subtask switching; action
generation remains closed-loop at every policy step.

\FloatBarrier

\section{Manipulation-Pattern Definition and Annotation}
\label{sec:supp_annotation}

\subsection{Operational Definition}
\begin{figure*}[!t]
\centering
\includegraphics[width=\textwidth]{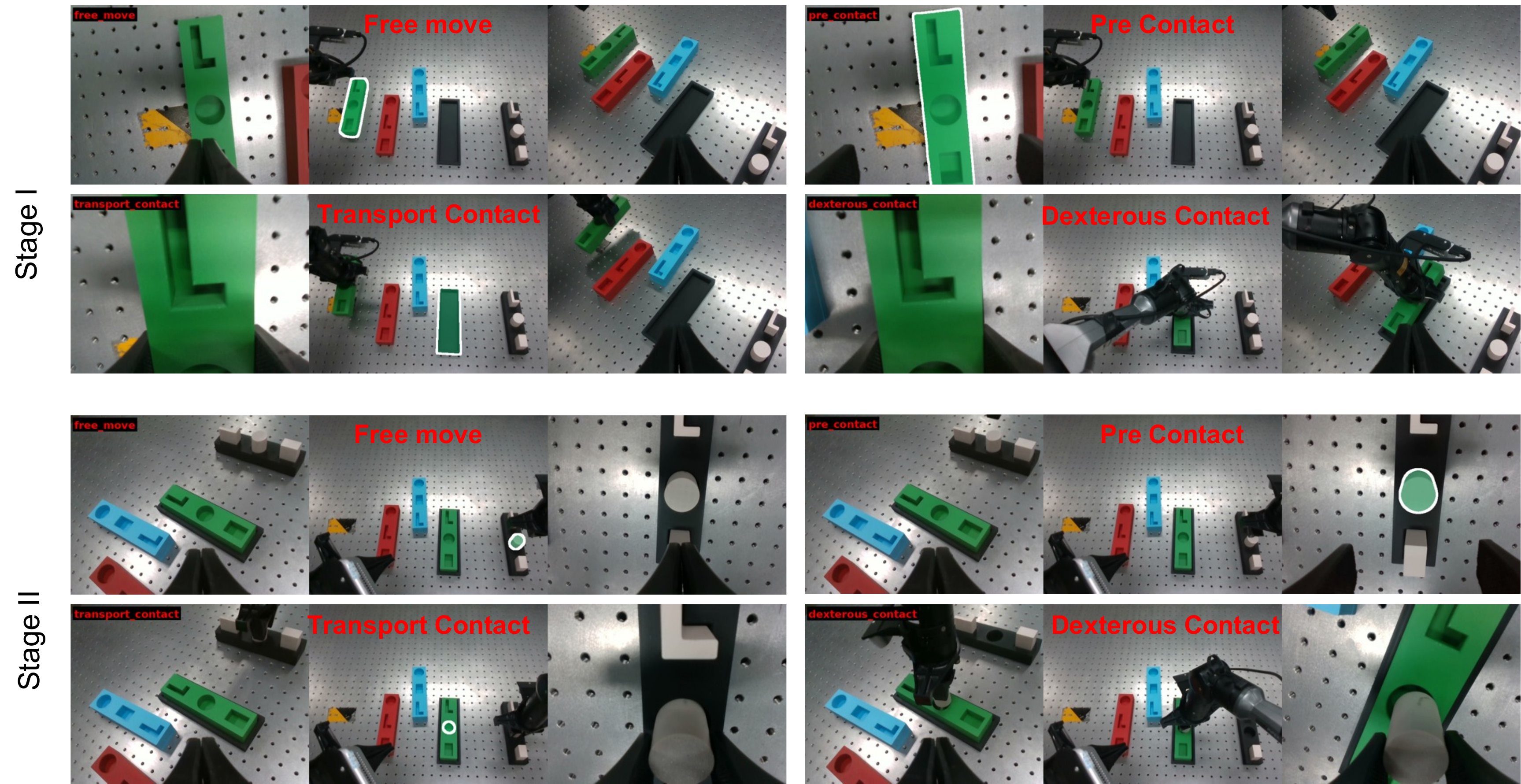}
\caption{Manipulation-pattern transitions in the two-stage peg-in-hole task.
Stage~I inserts the selected color or position base, and Stage~II inserts the corresponding shape peg.
Each pattern is illustrated with synchronized multi-view observations; the
outlined region denotes the active semantic target.}
\label{fig:pattern_transition}
\end{figure*}

Figure~\ref{fig:pattern_transition} instantiates the four interaction patterns
in the two-stage peg-in-hole task. Unlike a distance-threshold decomposition,
the annotation is anchored to observable interaction events---final approach,
stable grasp or contact, transport, release, and sustained fixture contact.
The same criteria therefore apply across objects, layouts, and task stages.

\textbf{Free move.} The gripper holds no task object and the arm moves without
sustained task contact. The segment ends at the final target-directed approach.
The global view is routed because scene-level target and goal geometry dominates
local contact geometry.

\textbf{Pre-contact.} The arm is in its final approach, with the gripper aligned
to grasp or establish contact. The active wrist view is routed for local
geometry. The segment ends at stable grasp or sustained contact; an incidental
one-frame touch does not define a boundary.

\textbf{Transport contact.} A stable grasp couples the object to the gripper
during motion toward its goal. The global view is routed to preserve the
object--goal relationship. The segment ends at release or at the onset of
contact-rich destination alignment.

\textbf{Dexterous contact.} The object remains in sustained contact with the
fixture, and completion requires local, force-informed correction. The active
wrist view and recent joint-effort history are therefore most informative. In
Fig.~\ref{fig:pattern_transition}, this pattern covers final base seating in
Stage~I and force-sensitive peg insertion in Stage~II.

For routing, \emph{pre-contact} and \emph{dexterous contact} require local wrist-view observations and are therefore resolved by the active wrist (left or right) on our bimanual platform, which has one wrist camera per arm. In contrast, \emph{free move} and \emph{transport contact} share the global route. Thus, the four task-independent manipulation-pattern families are expanded into six view-resolved routing labels: two global classes and four wrist-resolved classes. This expansion is specific to the multi-wrist observation setting; on a single-arm platform with a single wrist camera, the same four pattern families would correspond directly to four routing labels. Progress prediction remains defined at the pattern-family level, with one progress head for each of the four manipulation patterns.

\subsection{Boundary and Progress Annotation}
Annotators assign boundaries on synchronized multi-view video with
proprioception and joint effort available as supporting evidence. A contiguous
run with a constant six-class label defines one segment. For a segment spanning
frames $s,\ldots,e$, the normalized progress target is generated, rather than
annotated independently, as
\begin{equation}
\phi_t = \frac{t-s}{\max(e-s,1)}, \qquad s\leq t\leq e.
\label{eq:supp_progress}
\end{equation}
Hence, the first and last frames have progress 0 and 1 for segments longer than
one frame; a one-frame segment is assigned 0. Progress is recomputed after any
boundary correction, so no independent progress annotation is required.
\FloatBarrier

\section{Semantic Commitment, Grounding, and Tracking}
\label{sec:supp_reasoning}

This section gives the operational realization of the main paper's
``what-and-where'' decomposition. The Task Manager establishes a discrete
semantic commitment at manipulation-pattern events; the Semantic Grounder
localizes that commitment in the routed view; and visual tracking preserves its
image-space identity between events. Semantic identity is therefore updated
sparsely, while spatial state remains closed-loop.

\subsection{Task Manager and Execution Context}
The Task Manager converts a high-level instruction into sparse semantic
commitments synchronized with manipulation-pattern transitions. At reset,
Qwen3-VL-8B-Instruct~\cite{bai2025qwen3}, optionally supported by
open-vocabulary detection, interprets the instruction and scene as an ordered
plan
\begin{equation}
\mathcal{P}=\left[\sigma_r\right]_{r=1}^{K},
\label{eq:supp_plan}
\end{equation}
where $\sigma_r$ denotes the entities, goals, and relations associated with
subtask $r$. At the $k$th pattern event, the execution context is
\begin{equation}
\mathcal{C}_k=\big(\mathcal{P},\mathcal{H}_k,r_k\big),
\label{eq:supp_context}
\end{equation}
where $\mathcal{H}_k$ records completed subtasks and $r_k$ indexes the active
subtask. The Task Manager then resolves
\begin{equation}
(\ell_k,o_k^\star)
=
\mathcal{T}\!\left(
\ell,\hat{P}_{t_k},\mathcal{C}_k
\right),
\end{equation}
yielding the current instruction $\ell_k$ and semantic target $o_k^\star$.

The manipulation pattern determines the role of the target: free move and
pre-contact generally refer to the object being acquired, whereas transport
and dexterous contact shift attention toward its goal or interaction region.
This object--goal binding is preserved within the subtask and updated only
after the corresponding pattern cycle is completed. The resulting event-driven
commitment prevents transient perceptual changes from repeatedly redefining
task intent. In the reported tasks, it supports category--receptacle binding in
sorting, ordered letter selection in spelling, and block-to-slot followed by
peg-to-hole execution in insertion.

The following representative outputs document how the semantic plan is
instantiated in the reported experiments.

\begin{lstlisting}[style=prompt]
Sorting object:
{"label":"<short lowercase object name>",
 "category":"fruit|vegetable"}

Spelling target:
<one lowercase English word>

Peg-in-hole plan:
{"selected_block_color":"<color>",
 "selected_block_holes_top_to_bottom":["<shape>",...],
 "selected_peg_color":"<color>",
 "selected_peg_shape":"l shaped|circular|rectangular",
 "instruction":"<canonical two-step instruction>"}
\end{lstlisting}

At each grounding event, the active entity is expressed as a concise target
phrase. Direct templates are used for unambiguous entities, while Qwen resolves
more complex attributes or relations. The routed image and target phrase are
then supplied to the grounding model using the following prompt:

\begin{lstlisting}[style=prompt]
Locate exactly one visible object or region described as:
'<TARGET_PHRASE>'. Match the description as closely as possible.
If partially visible or occluded, return a tight box around the best
matching individual target. Do not box its container, holder, or a
group of similar objects. Output JSON in the form
[{"bbox_2d":[x1,y1,x2,y2]}], using coordinates normalized to
0--1000. Output [] when no matching target is visible.
\end{lstlisting}

\subsection{Semantic grounder}
Grounding DINO (Swin-T)~\cite{liu2024groundingdino} is used for efficient
object-level inventory when category semantics are direct; Qwen is used when
fine-grained attributes, spatial relations, or compositional disambiguation are
required. In the reported configuration, DINO initializes the sorting and
spelling inventories at reset, whereas Qwen parses the peg scene and performs
all transition-triggered phrase grounding. Thus the DINO and Qwen latency rows
in the main paper characterize two available grounding regimes; they are not
both invoked at every transition. Previously resolved boxes and fixed geometric
regions are reused when their semantic identity remains valid.

\begin{table}[H]
\centering
\caption{Grounding and tracking settings in the reported experiments. Box and
text thresholds apply to the reset-time DINO detector; ``tokens'' is the maximum
Qwen coordinate-output budget.}
\label{tab:grounder_config}
\scriptsize
\setlength{\tabcolsep}{2.7pt}
\resizebox{\columnwidth}{!}{%
\begin{tabular}{@{}lccc@{}}
\toprule
Item & Sorting & Spelling & Peg insertion \\
\midrule
Reset parser / detector & DINO & DINO & Qwen \\
Transition grounder & Qwen & Qwen & Qwen \\
Qwen box tokens & 16 & 16 & 64 \\
DINO box/text threshold & 0.22/0.18 & 0.25/0.20 & n/a \\
SAM2.1 checkpoint & Hiera-S & Hiera-S & Hiera-S \\
Tracker memory window & 24 & 24 & 24 \\
\bottomrule
\end{tabular}
}
\end{table}

\subsection{Goal-Consistent Tracking and Re-grounding}
\label{sec:supp_tracking}
Each camera maintains an independent state for the official memory-based video
propagation interface of SAM2.1 Hiera-S~\cite{ravi2025sam}. A grounding box
initializes a dense target mask, which is then propagated with a 24-frame memory
window. Only views selected by the current manipulation pattern are updated;
other memories are retained until those cameras are routed again. This implements
the main paper's distinction between a fixed semantic commitment and its
time-varying spatial realization.

When tracking confidence falls below $\tau_{\mathrm{trk}}=0.5$, the same
committed target phrase is sent back to the Semantic Grounder and the returned
box reinitializes that camera's memory. A target-phrase change always starts a
new tracker state. A pattern transition also reinitializes tracking when it
changes the routed view or the active entity; otherwise the existing commitment
is preserved. Consequently, re-grounding repairs spatial drift without silently
changing semantic identity.

\FloatBarrier

\section{Experiment Details}
\label{sec:supp_experiments}

The supplementary analysis follows the four questions in the main paper. It
adds trial-level OOD layouts and score accounting for Q1--Q2, clarifies the
routing and latency accounting used for Q3--Q4, and decomposes the remaining
errors without restating the headline comparisons.

\subsection{Metrics, Results, and Statistical Reporting}
We evaluate performance at three complementary levels. Let
$y_i\in\{0,1\}$ indicate whether the $i$th semantic decision selects the
correct target, $s_j\in\{0,0.5,1\}$ denote the credited outcome of the $j$th
required subtask, and $z_r\in\{0,1\}$ indicate whether trial $r$ is completed
successfully. The three metrics are
\begin{align}
\mathrm{IS}
&=
\frac{1}{N_{\mathrm{IS}}}
\sum_{i=1}^{N_{\mathrm{IS}}} y_i,
\\
\mathrm{Sub.\,SR}
&=
\frac{1}{N_{\mathrm{sub}}}
\sum_{j=1}^{N_{\mathrm{sub}}} s_j,
\\
\mathrm{Full\,SR}
&=
\frac{1}{N_{\mathrm{trial}}}
\sum_{r=1}^{N_{\mathrm{trial}}} z_r.
\end{align}
IS measures whether the correct semantic entity is selected, Sub.\ SR measures
the execution outcome of each required manipulation, and Full SR requires the
entire ordered task to be completed without failure.

\begin{figure*}[!t]
\centering
\includegraphics[width=0.94\textwidth]{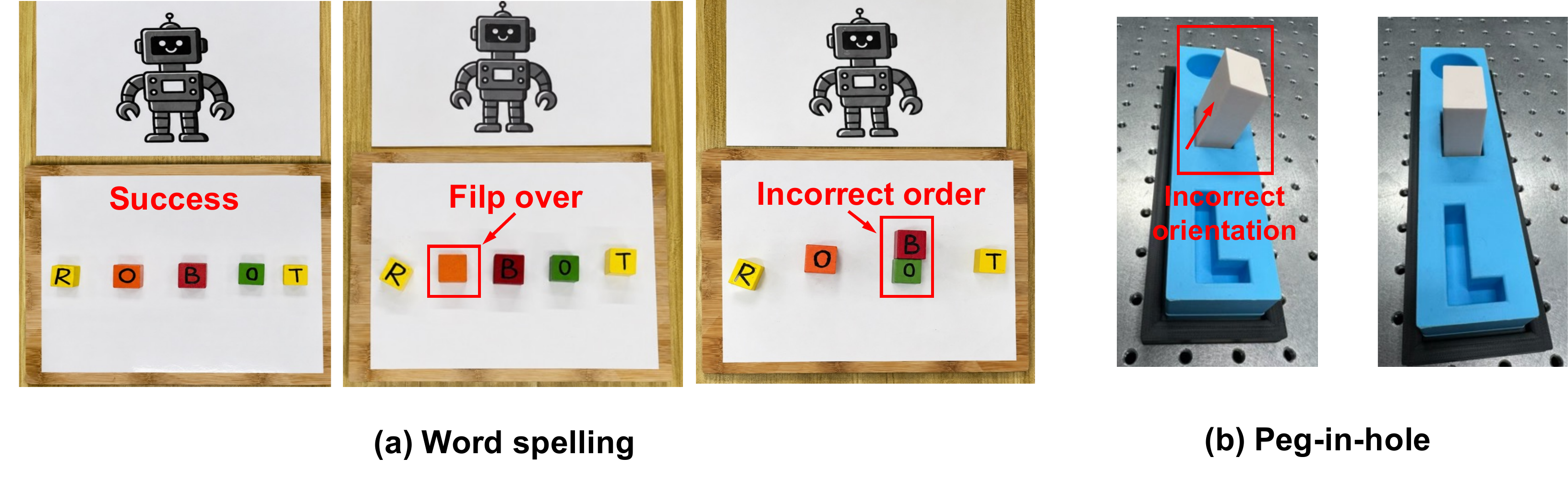}
\caption{Subtask-level scoring criteria. A correctly placed spelling block or
correctly aligned peg block receives a score of 1.0. A face-down letter, an
incorrect letter position, or a correctly transferred peg block with incorrect
final orientation receives 0.5. All partial outcomes receive zero credit under
Full SR.}
\label{fig:partial_credit}
\end{figure*}

Partial credit is used only for Sub.\ SR. Sorting subtasks are scored
binary. In spelling, selecting the correct letter but placing it face-down or
at an incorrect sequence position receives 0.5. In peg insertion, transferring
the correct block with an incorrect final orientation receives 0.5. IS remains
binary, and any partially completed subtask causes the corresponding trial to
fail under Full SR. Figure~\ref{fig:partial_credit} illustrates these scoring
criteria.

Table~\ref{tab:event_results} reports both the total number of evaluated
opportunities and the corresponding success rate. The ID/OOD evaluations
contain 113/46 sorting subtasks, 47/44 spelling subtasks, and 40/20
peg-insertion subtasks, respectively; the exact realized denominator for each
method is shown in the table. Peg insertion contains more semantic decisions
than manipulation subtasks because object, fixture, and insertion-target
selection are evaluated separately by IS.

\begin{table}[H]
\centering
\caption{Event-level semantic and subtask results. Each entry reports credited
successes over the total number of opportunities, followed by the corresponding
success rate in parentheses. Half-integer credits occur only in Sub.\ SR.
``Wall'' denotes Wall-OSS-0.5.}
\label{tab:event_results}
\scriptsize
\setlength{\tabcolsep}{2.4pt}
\resizebox{\columnwidth}{!}{%
\begin{tabular}{@{}llccc@{}}
\toprule
Method & Metric & Sorting & Spelling & Peg insertion \\
\midrule
\multicolumn{5}{@{}l}{\textit{In-distribution}} \\
\addlinespace[1pt]
\multirow{2}{*}{Wall}
& IS       & 60/113 (53.1\%)  & 21/47 (44.7\%) & 39/60 (65.0\%) \\
& \subsr{} & 60/113 (53.1\%)  & 17/47 (36.2\%) & 10/40 (25.0\%) \\
\addlinespace[1pt]
\multirow{2}{*}{\pis{}}
& IS       & 59/113 (52.2\%)  & 21/47 (44.7\%) & 21/60 (35.0\%) \\
& \subsr{} & 59/113 (52.2\%)  & 20/47 (42.6\%) & 11/40 (27.5\%) \\
\addlinespace[1pt]
\multirow{2}{*}{Wall+\ours{}}
& IS       & 97/113 (85.8\%)  & 39/47 (83.0\%) & 52/60 (86.7\%) \\
& \subsr{} & 97/113 (85.8\%)  & 38/47 (80.9\%) & 19/40 (47.5\%) \\
\addlinespace[1pt]
\multirow{2}{*}{\pis{}+\ours{}}
& IS       & 103/113 (91.2\%) & 46/47 (97.9\%) & 60/60 (100.0\%) \\
& \subsr{} & 103/113 (91.2\%) & 45/47 (95.7\%) & 27/40 (67.5\%) \\
\midrule
\multicolumn{5}{@{}l}{\textit{Out-of-distribution}} \\
\addlinespace[1pt]
\multirow{2}{*}{Wall}
& IS       & 22/46 (47.8\%) & 16/44 (36.4\%)   & 15/30 (50.0\%) \\
& \subsr{} & 17/46 (37.0\%) & 14/44 (31.8\%)   & 5/20 (25.0\%) \\
\addlinespace[1pt]
\multirow{2}{*}{\pis{}}
& IS       & 26/46 (56.5\%) & 18/44 (40.9\%)   & 8/30 (26.7\%) \\
& \subsr{} & 24/46 (52.2\%) & 16/44 (36.4\%)   & 4/20 (20.0\%) \\
\addlinespace[1pt]
\multirow{2}{*}{Wall+\ours{}}
& IS       & 34/46 (73.9\%) & 36/44 (81.8\%)   & 25/30 (83.3\%) \\
& \subsr{} & 30/46 (65.2\%) & 32/44 (72.7\%)   & 6/20 (30.0\%) \\
\addlinespace[1pt]
\multirow{2}{*}{\pis{}+\ours{}}
& IS       & 38/46 (82.6\%) & 42/44 (95.5\%)   & 27/30 (90.0\%) \\
& \subsr{} & 37/46 (80.4\%) & 38.5/44 (87.5\%) & 11/20 (55.0\%) \\
\bottomrule
\end{tabular}
}
\end{table}

Across both backbones and evaluation splits, HINT consistently improves IS and
Sub.\ SR. For \pis{}+\ours{}, the close alignment between IS and Sub.\ SR in
sorting and spelling indicates that correct semantic commitments are reliably
translated into executable actions. Peg insertion presents a different regime:
high ID/OOD IS (100.0\%/90.0\%) but lower Sub.\ SR (67.5\%/55.0\%) identifies
geometric precision and contact-sensitive control, rather than semantic
binding, as the remaining bottleneck.

Table~\ref{tab:fullsr_ci} reports trial-level Full SR with two-sided 95\%
Wilson confidence intervals. The ID evaluation contains 20 sorting trials,
15 spelling trials, and 20 peg-insertion trials; each OOD condition contains
10 trials. Wilson intervals are used because they remain well-defined at
observed rates of 0\% and 100\%, unlike standard Wald intervals.

\begin{table}[H]
\centering
\caption{Full-task success rate with two-sided 95\% confidence intervals.
Entries are reported as percentage $[\mathrm{lower},\mathrm{upper}]$ using
Wilson score intervals. ``Wall'' denotes Wall-OSS-0.5.}
\label{tab:fullsr_ci}
\scriptsize
\setlength{\tabcolsep}{2.8pt}
\resizebox{\columnwidth}{!}{%
\begin{tabular}{@{}lccc@{}}
\toprule
Method & Sorting & Spelling & Peg insertion \\
\midrule
\multicolumn{4}{@{}l}{\textit{In-distribution}} \\
Wall
& $0.0\ [0.0,16.1]$
& $13.3\ [3.7,37.9]$
& $5.0\ [0.9,23.6]$ \\
\pis{}
& $10.0\ [2.8,30.1]$
& $13.3\ [3.7,37.9]$
& $5.0\ [0.9,23.6]$ \\
Wall+\ours{}
& $40.0\ [21.9,61.3]$
& $40.0\ [19.8,64.3]$
& $10.0\ [2.8,30.1]$ \\
\pis{}+\ours{}
& $60.0\ [38.7,78.1]$
& $86.7\ [62.1,96.3]$
& $40.0\ [21.9,61.3]$ \\
\midrule
\multicolumn{4}{@{}l}{\textit{Out-of-distribution}} \\
Wall
& $0.0\ [0.0,27.8]$
& $0.0\ [0.0,27.8]$
& $10.0\ [1.8,40.4]$ \\
\pis{}
& $0.0\ [0.0,27.8]$
& $0.0\ [0.0,27.8]$
& $0.0\ [0.0,27.8]$ \\
Wall+\ours{}
& $20.0\ [5.7,51.0]$
& $20.0\ [5.7,51.0]$
& $10.0\ [1.8,40.4]$ \\
\pis{}+\ours{}
& $30.0\ [10.8,60.3]$
& $30.0\ [10.8,60.3]$
& $30.0\ [10.8,60.3]$ \\
\bottomrule
\end{tabular}
}
\end{table}

The confidence intervals reflect the limited number of full-task trials and
should therefore be interpreted as uncertainty ranges rather than precise
pairwise rankings. In particular, the OOD estimates have wider intervals
because each condition contains only 10 trials. Nevertheless, the consistent
event-level improvements across tasks, splits, and action backbones provide
denser evidence that HINT improves semantic target selection and subtask
execution, while the Full SR results demonstrate that these local gains
translate into complete long-horizon task success.

\subsection{Semantic-Compositional OOD Variations}
We organize the OOD evaluation along four primary axes of semantic variation:

\begin{itemize}
\item \textbf{Entity and Category Variation.}
Sorting introduces previously unseen produce categories, while spelling changes
the target letters and landmark concepts. Peg insertion varies the candidate
blocks and pegs presented in the scene. These settings test whether the system
can identify task-relevant entities beyond the specific semantic instances
observed during training.

\item \textbf{Visual-Attribute and Goal Variation.}
The evaluation changes object colors, goal appearances, and attribute
combinations. Sorting replaces familiar receptacles with visually distinct
goals; spelling introduces new letter colors and visual configurations; and peg
insertion uses novel color--shape compositions. These variations require the
system to resolve target identity without relying on fixed appearance
correspondences.

\item \textbf{Spatial and Relational Variation.}
Object arrangements, distractor layouts, and candidate ordering are varied
across scenes. Peg insertion further introduces relational descriptions such as
\emph{left}, \emph{right}, \emph{above}, and \emph{below}. The intended target
must therefore be determined from its relation to other scene elements rather
than from a memorized location.

\item \textbf{Instruction-Conditioned Composition.}
The instruction changes how entities, attributes, and goals should be composed.
Examples include altered category--goal assignments in sorting, unseen target
words and letter sequences in spelling, and new block--slot and peg--hole
bindings in insertion. This axis evaluates whether familiar semantic elements
can be recombined into previously unseen task specifications.
\end{itemize}

\begin{figure*}[!t]
\centering
\includegraphics[width=0.99\textwidth]{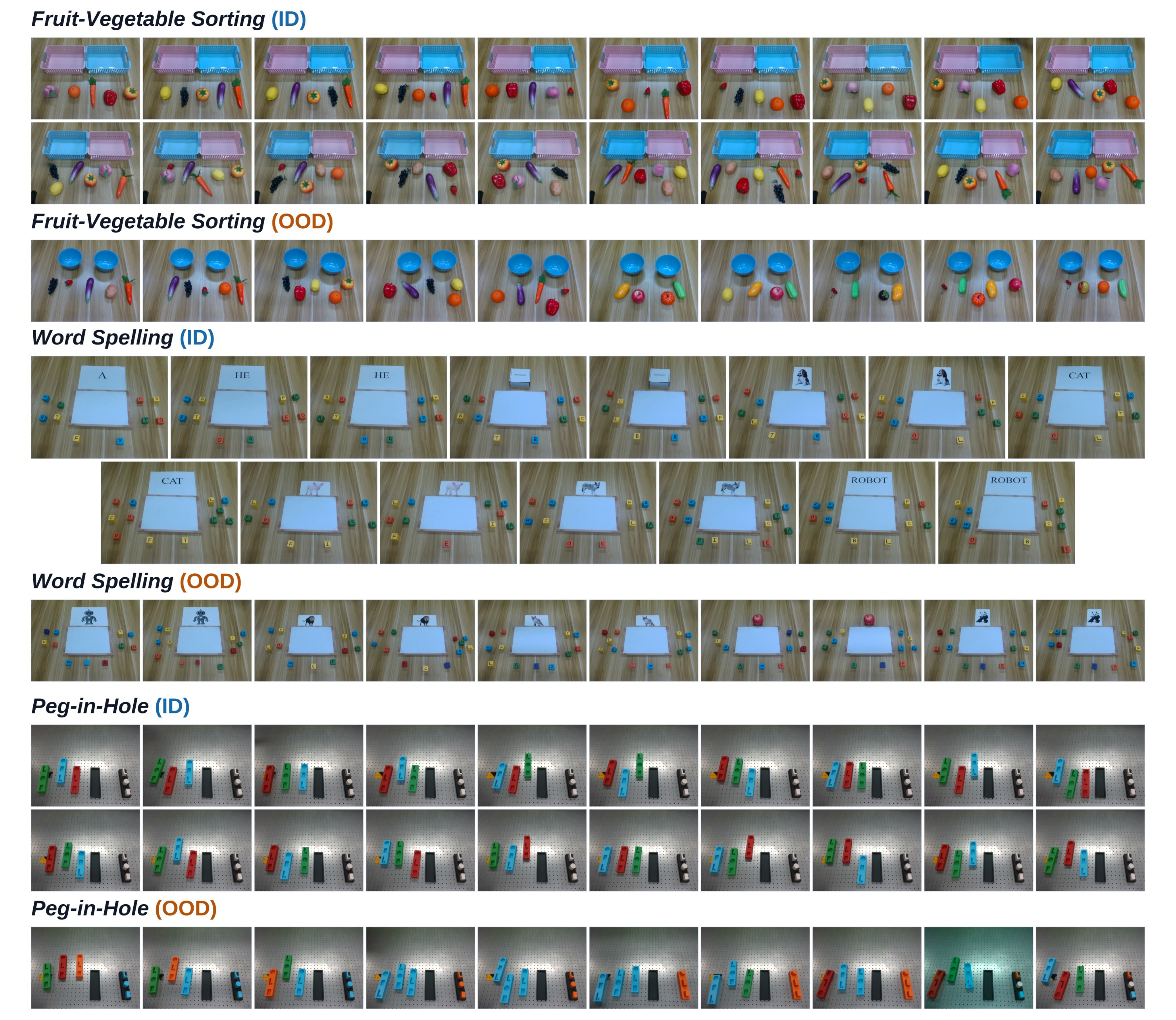}
\caption{Complete deployed ID and OOD layouts for sorting (top), spelling
(middle), and peg-in-hole insertion (bottom), illustrating variations in entity
identity, visual attributes, spatial relations, and instruction-conditioned
semantic composition.}
\label{fig:eval_scenes}
\end{figure*}

Figure~\ref{fig:eval_scenes} presents the complete deployed ID and OOD layouts.
The montages show how these variation axes are instantiated and combined within
each task.

\FloatBarrier


\subsection{Routing and Latency Accounting}
The two routing measures in the main paper answer different parts of Q3.
Pattern prediction evaluates the shared interaction-state representation,
whereas Switch SR evaluates the downstream event interface: a switch is correct
only when a detected transition activates the intended next semantic target.
The reported 251/260 successful switches decompose as 96/100 for sorting,
95/100 for spelling, and 60/60 for insertion. The high switch rate despite lower
pattern reliability near insertion boundaries is consistent with the
progress gate in Sec.~\ref{sec:supp_impl}: stable semantic events need not
require every boundary-adjacent frame to be classified correctly.

For Q4, the aggregate frame times in the main paper are the sums of the modules
on the corresponding execution path. With the reported component times, a DINO
reasoning frame is $185+14+227=426$~ms, a Qwen reasoning frame is
$185+14+431=630$~ms, and a tracking frame is $185+14+43=242$~ms. Excluding the
185-ms action policy, the auxiliary cost is therefore 241/445~ms for a
DINO/Qwen reasoning event and 57~ms for a tracking step. This accounting makes
the source of the reported efficiency gain explicit: HINT reduces the frequency
of semantic reasoning rather than weakening the reasoning model or opening the
action loop.

\subsection{Semantic-Interface Ablation}
The OOD spelling ablation uses the same 44 semantic/subtask opportunities and
10 trials for every variant. Attention-only (without highlighting) obtains IS
$26/44$, \subsr{} $25.5/44$, and \fullsr{} $1/10$; highlighting-only (without
the attention prior) obtains $39/44$, $32.5/44$, and $1/10$; the combined
interface obtains $42/44$, $38.5/44$, and $3/10$. These counts complement the
main-paper percentages and show that the combined gain is present in both
binary target selection and credited execution, not only in the strict
full-task endpoint.

\subsection{Failure Analysis}
Each unsuccessful trial is assigned to its earliest irreversible error, avoiding
the repeated attribution of one upstream mistake to multiple downstream stages.
\emph{Reason} denotes incorrect subtask inference or target grounding;
\emph{Pick} denotes failure to acquire the correct object; \emph{Place} covers
transport, ordering, release, or orientation errors; and \emph{Contact} denotes
force-sensitive alignment or insertion failure. These categories are mutually
exclusive.

\textbf{Baseline failures.}
The baselines fail primarily because semantic knowledge is not reliably
translated into target-conditioned action. This is most evident in sorting:
the base \pis{} achieves 89.5\% Sub.\ SR when the familiar visual mapping remains
valid, but drops to 52.2\% when the instruction changes the category--goal
assignment. The policy retains the required grasp-and-transport capability, yet
defaults to learned visual correspondences when they conflict with language.
The low OOD IS values and zero Full SR of the base \pis{} across all three tasks
reflect the same failure at longer horizons: an early target-selection error
invalidates the remaining action sequence. Peg insertion additionally exposes
an action-side limitation, as correct target selection does not guarantee the
precision and contact correction required for successful insertion.

\textbf{Failures with HINT.}
HINT externalizes the selected target as a persistent spatial referent and
therefore removes many upstream binding failures. The remaining errors are
summarized in Table~\ref{tab:failure_counts}.

The residual failure distribution reveals task-dependent bottlenecks. Sorting
failures remain entirely semantic, particularly under OOD conditions, indicating
that unseen categories or ambiguous category--goal mappings can still exceed
the Task Manager or grounder's open-world reasoning capacity. Spelling exhibits
a mixed regime: HINT largely resolves the intended letter sequence, but dense
distractors and fine-grained letter appearances can still produce grounding,
grasping, or placement errors. In peg insertion, reasoning is never the first
cause; 11 of 12 ID failures and all seven OOD failures occur during placement or
contact-rich insertion. Here HINT selects the correct target, but the unchanged
action backbone remains limited by geometric precision and force-sensitive
correction.

\begin{table}[H]
\centering
\caption{First-cause trial outcomes for \pis{}+\ours{}. Categories are mutually
exclusive and sum to the number of trials. ``Place'' includes transport and
placement errors before contact-rich insertion; ``--'' indicates that the
contact category is not applicable.}
\label{tab:failure_counts}
\scriptsize
\setlength{\tabcolsep}{2.8pt}
\begin{tabular}{@{}llrrrrr@{}}
\toprule
Split & Task & Success & Reason & Pick & Place & Contact \\
\midrule
\multirow{3}{*}{ID}
& Sorting       & 12/20 & 8 & 0 & 0 & -- \\
& Spelling      & 13/15 & 1 & 1 & 0 & -- \\
& Peg insertion & 8/20  & 0 & 1 & 4 & 7  \\
\midrule
\multirow{3}{*}{OOD}
& Sorting       & 3/10 & 7 & 0 & 0 & -- \\
& Spelling      & 3/10 & 2 & 2 & 3 & -- \\
& Peg insertion & 3/10 & 0 & 0 & 4 & 3  \\
\bottomrule
\end{tabular}
\end{table}

This migration of failures is more informative than a uniform reduction in
error counts. HINT primarily addresses the semantic interface between reasoning
and action; once this bottleneck is removed, the dominant errors shift toward
the next weakest component. Residual reasoning errors therefore define the
limit of open-world semantic interpretation, whereas placement and contact
errors expose the control ceiling of the underlying action policy.